\documentclass[sigconf,nonacm]{acmart}

\usepackage[ruled,vlined,linesnumbered, noend]{algorithm2e}
\usepackage[lambda,advantage,operators,sets,adversary,landau,probability,notions,logic,ff,mm,primitives,events,complexity,asymptotics,keys]{cryptocode}
\usepackage{booktabs}
\usepackage{threeparttable}

\usepackage{caption}
\usepackage{subcaption}

\newcommand{\tsum}{\textstyle{\sum}}

\AtBeginDocument{%
  \providecommand\BibTeX{{%
    \normalfont B\kern-0.5em{\scshape i\kern-0.25em b}\kern-0.8em\TeX}}}

\setcopyright{none}
\copyrightyear{2021}
\acmYear{2021}

\begin{document}

\title{FedV: Privacy-Preserving Federated Learning over Vertically Partitioned Data}

\author{Runhua Xu$\dagger$, Nathalie Baracaldo$\dagger$, Yi Zhou$\dagger$, Ali Anwar $\dagger$, James Joshi$\ddagger$, Heiko Ludwig$\dagger$}
\email{
{runhua,yi.zhou,ali.anwar2}@ibm.com, {baracald,hludwig}@us.ibm.com, jjoshi@pitt.edu
}
\affiliation{%
  \institution{$\dagger$ IBM Research, San Jose, CA, USA}
}
\affiliation{%
  \institution{$\ddagger$ University of Pittsburgh, Pittsburgh, PA, USA}
}

\renewcommand{\shortauthors}{Xu, et al.}

\begin{abstract}
Federated learning (FL) has been proposed to allow collaborative training of machine learning (ML) models among multiple parties where each party can keep its data private. In this paradigm, only \textit{model updates}, such as model weights or gradients, are shared. 
Many existing approaches have focused on \textit{horizontal} FL, where each party has the entire feature set and labels in the training data set.
However, many real scenarios follow a \textit{vertically-partitioned} FL setup,
where a complete feature set is formed only when all the datasets from the parties are combined, and the labels are only available to a single party.
Privacy-preserving \textit{vertical FL} is challenging because complete sets of labels and features are not owned by one entity.
Existing approaches for \textit{vertical FL} require multiple peer-to-peer communications among parties, leading to lengthy training times, and are restricted to (approximated) linear models and just two parties.
To close this gap, we propose \textit{FedV}, a framework for secure gradient computation in vertical settings for several widely used ML models such as linear models, logistic regression, and support vector machines.
\textit{FedV} removes the need for peer-to-peer communication among parties by using functional encryption schemes; this allows \textit{FedV} to achieve faster training times.
It also works for larger and changing sets of parties.
We empirically demonstrate the applicability for multiple types of ML models and show a reduction of 10\%-70\% of training time and 80\% to 90\% in data transfer with respect to the state-of-the-art approaches.
\end{abstract}



\settopmatter{printfolios=true}

\maketitle

\section{Introduction}

Machine learning (ML) has become ubiquitous and instrumental in many applications such as predictive maintenance, recommendation systems, self-driving vehicles, and healthcare.
The creation of ML models requires training data that is often subject to privacy or regulatory constraints, restricting the way data can be shared, used and transmitted.
Examples of such regulations include the European General Data Protection Regulation (GDPR), California Consumer Privacy Act (CCPA) and Health Insurance Portability and Accountability Act (HIPAA), among others.

There is great benefit in building a predictive ML model over datasets from multiple sources. This is because a single entity, henceforth referred to as a party, may not have enough data to build an accurate ML model. However, regulatory requirements and privacy concerns may make pooling such data from multiple sources infeasible.
\textit{Federated learning} (FL) \cite{mcmahan2016communication, konevcny2016federated} has recently been shown to be very promising for enabling a collaborative training of models among multiple parties - under the orchestration of an \textit{aggregator} - without having to share any of their raw training data.
In this paradigm, only \textit{model updates}, such as model weights or gradients, need to be exchanged.

There are two types of FL approaches, \textit{horizontal and vertical FL}, which mainly differ in the data available to each party.
In \textit{horizontal FL}, each party has access to the entire feature set and labels; thus, each party can train its local model based on its own dataset. All the parties then share their model updates with an aggregator and the aggregator then creates a global model by combining, e.g., averaging, the model weights received from individual parties.
In contrast, \textit{vertical FL} (VFL) refers to collaborative scenarios where individual parties do \textit{not} have the complete set of features and labels  and, therefore, cannot train a model using their own datasets locally.
In particular, parties' datasets need to be aligned to create the complete feature vector without exposing their respective training data, and the model training needs to be done in a privacy-preserving way.

\begin{table*}
  \caption{Comprehensive Comparison of Emerging VFL Solutions}
  \label{tab:compare}
  \vspace{-2mm}
  \footnotesize
  \begin{threeparttable}
  \begin{tabular}{ccccl}
    \toprule
    Proposal & Communication $\dagger$ & Computation & Privacy-Preserving Approach & Supported Models with SGD training\\
    \midrule
    Gasc{\'o}n et al. \cite{gascon2016secure} & \textit{mpc} + 1 round p2c & garbled circuits & hybrid MPC  & linear regression \\
    Hardy et al. \cite{hardy2017private} & \textit{p2p} + 1 round p2c & ciphertext & cryptosystem (partially HE) & logistic regression (LR) with Taylor approximation \\
    Yang et al. \cite{yang2019quasi} & \textit{p2p} + 1 round p2c & ciphertext  & cryptosystem (partially HE) & Taylor approximation based LR with quasi-Newton method \\
    Gu et al. \cite{gu2020federated}  & \textit{partial p2p} + 2 rounds p2c & normal  & random mask + tree-structured comm.  & non-linear learning with kernels \\
    Zhang et al. \cite{zhang2021secure} & \textit{partial p2p} + 2 rounds p2c & normal & random mask + tree-structured comm.  & logistic regression \\
    Chen et al. \cite{chen2020vafl}  & 2 rounds p2c & normal & local Gaussian DP perturbation & DP noise injected LR and neural networks\\
    Wang et al. \cite{wang2020hybrid}  & 2 rounds p2c & normal & joint Gaussian DP perturbation & DP noise injected LR\\
    \textit{FedV} (our work) & 1 round p2c & ciphertext & cryptosystem (Functional Encryption) & linear models, LR, SVM with kernels\\
    \bottomrule
    \end{tabular}
  \begin{tablenotes}
  \item[$\dagger$] The communication represents interaction topology needed per training epoch. Here, `p2p' presents the peer-to-peer communication among parties; `p2c' denotes the communication between each party and the coordinator (a.k.a, active party in some solutions); `mpc' indicates the extra communication required by the garbled circuits multi-party computation, e.g., oblivious transfer interactions. 
  \end{tablenotes}
   \end{threeparttable}
   \vspace{-3mm}
\end{table*}

Existing approaches, as shown in \tablename~\ref{tab:compare}, to train ML models in vertical FL or vertical setting,
are model-specific and rely on general (garbled circuit based) secure multi-party computation (SMC), differential privacy noise perturbation, or partially additive homomorphic encryption (HE) (i.e., Paillier cryptosystem \cite{damgaard2001generalisation}).
These approaches have several limitations:
First, they apply only to limited models. They require the use of Taylor series approximation to train non-linear ML models, such as logistic regression, that possibly reduces the model performance and \textit{cannot} be generalized to solve classification problems. Furthermore, the prediction and inference phases of these vertical FL solutions rely on approximation-based secure computation or noise perturbation. As such, these solutions cannot predict as accurately as a centralized ML model can.
Secondly, using such cryptosystems as part of the training process substantially increases the training time.
Thirdly, these protocols require a large number of peer-to-peer communication rounds among parties, 
making it difficult to deploy them in systems that have poor connectivity or
where communication is limited to a few specific entities due to regulation such as HIPAA.
Finally, other approaches such as the one proposed in \cite{yu2006privacy} require sharing class distributions, which may lead to potential leakage of private information of each party.


To address these limitations, we propose \textit{FedV}. This framework  substantially
reduces the amount of communication required to train ML models in a vertical FL setting. 
\textit{FedV} does not require any peer-to-peer communication among parties and can work with gradient-based training algorithms, such as stochastic gradient descent and its variants, to train a variety of ML models, e.g., logistic regression, support vector machine (SVM), etc.
To achieve these benefits, \textit{FedV} orchestrates multiple functional encryption techniques \cite{abdalla2015simple, abdalla2018multi} - which are non-interactive in nature - speeding up the training process compared to the state-of-the-art approaches.
Additionally, \textit{FedV} supports more than two parties and allows parties to dynamically leave and re-join without a need for re-keying.
This feature is not provided by garbled-circuit or HE based techniques utilized by state-of-the-art approaches.

To the best of our knowledge, this is the first generic and efficient privacy-preserving vertical federated learning (VFL) framework that
drastically reduces the number of communication rounds required during model training while supporting a wide range of widely used ML models. The main \textbf{\textit{contributions}} of this paper are as follows: 

We propose \textit{FedV}, a generic and efficient privacy-preserving vertical FL framework, which only requires communication between parties and the aggregator as a one-way interaction and does not need any peer-to-peer communication among parties.

\textit{FedV} enables the creation of highly accurate models as it does not require the use of Taylor series approximation to address non-linear ML models.
In particular, \textit{FedV} supports stochastic gradient-based algorithms to train many classical ML models,
such as, linear regression, logistic regression and support vector machines, among others, without requiring linear approximation for nonlinear ML objectives as a mandatory step, as in the existing solutions. 
\textit{FedV} supports both lossless training and lossless prediction.


We have implemented and evaluated the performance of \textit{FedV}. Our results show that compared to existing approaches \textit{FedV} achieves 
significant improvements both in training time and communication cost without compromising privacy.
We show that these results hold for a range of widely used ML models including linear regression, logistic regression and support vector machines. 
Our experimental results show a reduction of 10\%-70\% of training time and 80\%-90\% of data transfer when compared to state-of-the art approaches.


\vspace{-2mm}
\section{Background}
\label{sec:bg}

\subsection{Vertical Federated Learning}

\label{sec:bg:vfl}

VFL is a powerful approach that can help create ML models for many real-world problems where a single entity does not have access to all the training features or labels.
Consider a set of banks and a regulator.
These banks may want to collaboratively create an ML model using their datasets to flag accounts involved in money laundering.
Such a collaboration is important as criminals typically use multiple banks to avoid detection.
However, if several banks join together to find a common vector for each client and a regulator provides the labels, showing which clients have committed money laundering, such fraud can be identified and mitigated.
However, each bank may not  want to share its clients' account details and in some cases it is even prevented to do so. 

One of the requirements for privacy-preserving VFL is thus to ensure that the dataset of each party are kept confidential.
VFL requires two different processes: \textit{entity resolution} and \textit{vertical training}.
Both of them are orchestrated by an \textit{Aggregator} that acts as a third semi-trusted party interacting with each party.
Before we present the detailed description of each process, we introduce the notation used throughout the rest of the paper.

\noindent \textbf{Notation:} 
Let $\mathcal{P} = \{p_i\}_{i\in[n]}$ be the set of $n$ parties 
in VFL. 
Let $\mathcal{D}^{[X,Y]}$ be the training dataset across the set of parties $\mathcal{P}$, where $X\in \mathbb{R}^d$ represents the feature set and $Y\in \mathbb{R}$ denotes the labels.
We assume that except for the identifier features, there are no overlapping \textit{training} features between any two parties' local datasets, and these datasets can form the ``global'' dataset $\mathcal{D}$.
As it is commonly done in VFL settings, we assume that only one party has the class labels, and we call it the \textit{active party}, while other parties are \textit{passive parties}.
For simplicity, in the rest of the paper, let $p_1$ be the \textit{active party}.
The goal of \textit{FedV} is to train a ML model $\mathcal{M}$ over the dataset $\mathcal{D}$ from the party set $\mathcal{P}$ without leaking any party's data.

\begin{figure}
    \centering
    \includegraphics[width=0.45\textwidth]{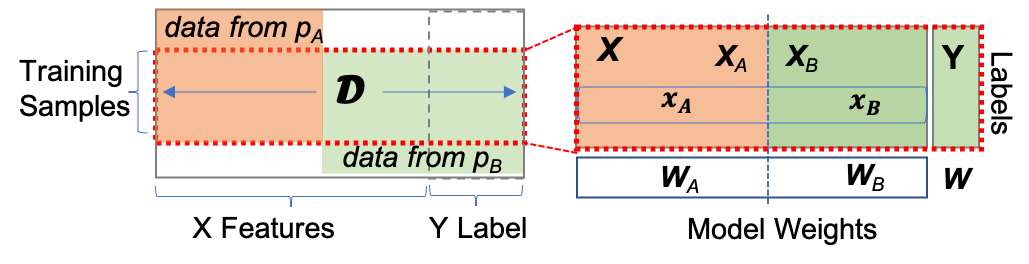}

    \caption{Vertically partitioned data across parties. In this example, $p_A$ and $p_B$ have overlapping identifier features, and $p_B$ is the active party that has the labels.}
    \label{fig:vfl}
    \vspace{-5mm}
\end{figure}

\noindent \textbf{Private Entity Resolution (PER):}
In VFL, unlike in a centralized ML scenario, $\mathcal{D}$ is distributed across multiple parties.
Before training takes place, it is necessary to \textit{`align'} the records of each party without revealing its data.
This process is known as entity resolution \cite{christen2012data}.
Figure~\ref{fig:vfl} presents a simple example of how $\mathcal{D}$ can be vertically partitioned among two parties.
After the entity resolution step, records from all parties are linked to form the complete set of \textit{training samples}.

Ensuring that the entity resolution process does not lead to inference of private data of each party is crucial in VFL.
A curious party should not be able to infer the presence or absence of a record.
Existing approaches, such as  \cite{nock2018entity, ion2019deploying}, use a bloom filter and random oblivious transfer \cite{dong2013private, kolesnikov2016efficient} with a shuffle process to perform private set intersection. This helps finding the matching record set while preserving privacy. 
We assume there exists shared {\sl record identifiers}, such as names, dates of birth or universal identification numbers, that can be used to perform entity matching.
In \textit{FedV}, we employ the anonymous linking code technique called \textit{cryptographic long-term key (CLK)} and matching method called \textit{Dice coefficient} \cite{schnell2011novel} to perform PER, as has been done in \cite{hardy2017private}.
As part of this process, each party generates a set of CLK based on the identifiers of the local dataset and shares it with the \textit{aggregator} that  matches the CLKs received and generate a permutation vector for each party to \textit{shuffle} its local dataset.
The shuffled local datasets are now ready to be used for private vertical training.





\noindent \textbf{Private Vertical Training:}
After the private entity resolution process takes place,
the training phase can start. This is the process this paper focuses on.
In the following, we discuss the basics of the gradient descent training process in detail.


\subsection{Gradient Descent in Vertical FL}

As the subsets of the feature set are distributed among different parties, gradient descent (GD)-based methods need to be adapted to such vertically partitioned settings.
We now explain how and why this process needs to be modified.
GD method \cite{nesterov1998introductory} represents a class of optimization algorithms that find the minimum of a target loss function; for example, in machine learning domain, a typical loss function can be defined as follows,
\begin{equation}
\label{equ:train_loss}
E_{\mathcal{D}}(\pmb{w}) = \tfrac{1}{n}\tsum^{n}_{i=1}\mathcal{L}(y^{(i)}, f(\pmb{x}^{(i)};\pmb{w})) + \lambda R(\pmb{w}),
\end{equation}
where $\mathcal{L}$ is the loss function, $y^{(i)}$ is the corresponding class label of data sample $\pmb{x}^{(i)}$, $\pmb{w}$ denotes the model parameters,  $f$ is the prediction function, and $R$ is regularization term with coefficient $\lambda$.
GD finds a solution of \eqref{equ:train_loss} by iteratively moving in the direction of the locally steepest descent as defined by the negative of the gradient, i.e., $\pmb{w} \leftarrow \pmb{w} - \alpha\nabla E_{\mathcal{D}}(\pmb{w}), $
where $\alpha$ is the learning rate, and $\nabla E_{\mathcal{D}}(\pmb{w})$ is the gradient computed at the current iteration.
Due to their simple algorithmic schemes, GD and its variants, like SGD, have become the common approaches to find the optimal parameters (a.k.a. the weights) of a ML model based on $\mathcal{D}$ \cite{nesterov1998introductory}.
In a VFL setting, since $\mathcal{D}$ is vertically partitioned among parties, the gradient computation $\nabla E_{\mathcal{D}}(\pmb{w})$ is more computationally involved than in a centralized ML setting.

Considering the simplest case where there are only two parties, $p_A, p_B$, in a VFL system as illustrated in Figure~\ref{fig:vfl}, and MSE (Mean Squared Loss) is used as the target loss function, i.e., $E_{\mathcal{D}}(\pmb{w}) = \frac{1}{n}\sum^{n}_{i=1}(y^{(i)}- f(\pmb{x}^{(i)};\pmb{w}))^2$, we have
\begin{align}
\nabla E_{\mathcal{D}}(\pmb{w}) 
&= -\tfrac{2}{n}\tsum^{n}_{i=1}(y^{(i)}- f(\pmb{x}^{(i)};\pmb{w}))\nabla f(\pmb{x}^{(i)};\pmb{w}).
\label{eq:gen_g_update}
\end{align}

If we expand \eqref{eq:gen_g_update} and compute the result of the summation, we need to compute $-y^{(i)}\nabla f(\pmb{x}^{(i)};\pmb{w})$ for $i=1,...n$, which requires feature information from both $p_A$ and $p_B$, and labels from $p_B$. And, clearly, $\nabla f(\pmb{x}^{(i)};\pmb{w})= [\partial_{\pmb{w}_A} f(\pmb{x}_A^{(i)};\pmb{w}); \partial_{\pmb{w}_B}f(\pmb{x}_B^{(i)};\pmb{w})]$ does not always hold for any function $f$, since $f$ may not be well-separable w.r.t. $\pmb{w}$. Even when it holds for linear functions like $f(\pmb{x}^{(i)};\pmb{w})= \pmb{x}^{(i)}\pmb{w}= \pmb{x}_A^{(i)}\pmb{w}_A + \pmb{x}_B^{(i)}\pmb{w}_B$, \eqref{eq:gen_g_update} will be reduced as follows:
\begin{align}
\nabla E_{\mathcal{D}}(\pmb{w}) 
&= -\tfrac{2}{n}\tsum^{n}_{i=1}(y^{(i)}- \pmb{x}^{(i)}\pmb{w})[\pmb{x}_A^{(i)};\pmb{x}_B^{(i)}]\nonumber\\
& = -\tfrac{2}{n}\tsum^{n}_{i=1}\left([(\textcolor{red}{y^{(i)}}-\pmb{x}_A^{(i)}\pmb{w}_A - \textcolor{red}{\pmb{x}_B^{(i)}\pmb{w}_B})\textcolor{red}{\pmb{x}_A^{(i)}};\right.\nonumber\\
&\quad \left.(y^{(i)}-\textcolor{red}{\pmb{x}_A^{(i)}\pmb{w}_A} - \pmb{x}_B^{(i)}\pmb{w}_B)\textcolor{red}{\pmb{x}_B^{(i)}}]\right),
\label{eq:linear_g_update}
\end{align}
This may lead to exposure of training data between two parties due to the computation of some terms (colored in red) in \eqref{eq:linear_g_update}.
Under the VFL setting, the gradient computation at each training epoch relies on (\romannumeral1) the parties' collaboration to exchange their ``partial model'' with each other, or (\romannumeral2) exposing their data to the aggregator to compute the final gradient update. 
Therefore, any naive solutions will lead to a significant risk of privacy leakage, which will counter the initial goal of 
the FL to protect data privacy.
Before presenting our approach, we first overview the basics of functional encryption.

\subsection{Functional Encryption}
\label{sec:bg:fe}

Our proposed \textit{FedV} makes use of encryption (FE) a cryptosystem that allows computing a specific
function over a set of ciphertexts without revealing the inputs. FE belongs to a public-key encryption family \cite{lewko2010fully, boneh2011functional}, 
where possessing a secret key called \textit{a functionally derived key} enables the computation of a function $f$ that takes as input ciphertexts, without revealing the ciphertexts.
The \textit{functionally derived key} is provided by a trusted third-party authority (TPA) which also responsible for initially setting up the cryptosystem.
For VFL, we require the computation of inner products.
For that reason, we adopt functional encryption for \textit{inner product} (FEIP), which allows the computation of the
inner product between two vectors $x$ containing encrypted private data, 
and $y$ containing public plaintext data.
To compute the inner product $\langle \pmb{x}, \pmb{y} \rangle$
the decrypting entity (e.g., aggregator) needs to obtain a \textit{functionally derived key} from the TPA. 
To produce this key, the TPA requires access to the \textit{public plaintext} vector $y$.
Note that the TPA does not need access to the private encrypted vector $x$.

We adopt two types of inner product FE schemes:
\textit{single-input functional encryption ($\mathcal{E}_\text{SIFE}$)} proposed in \cite{abdalla2015simple} and \textit{multi-input functional encryption ($\mathcal{E}_\text{MIFE}$)} introduced in \cite{abdalla2018multi}, which we explain in detail below.

\noindent\textbf{{SIFE}}($\mathcal{E}_\text{SIFE}$).
To explain this crypto system, consider the following simple example.
A party wants to keep $x$ private but wants an entity (aggregator) to be able to compute the inner product $\langle \pmb{x}, \pmb{y} \rangle$. 
Here $\pmb{x}$ is secret and encrypted and $\pmb{y}$ is public and provided by the aggregator to compute the inner product.
During set up, the TPA provides the public key $\texttt{pk}^{\text{SIFE}}$ to a party.
Then, the party encrypts $\pmb{x}$ with that key, denoted as
$ct_{\pmb{x}} = \mathcal{E}_\text{SIFE}.\text{Enc}_{\texttt{pk}^{\text{SIFE}}}(\pmb{x})$;
and sends $ct_{\pmb{x}}$ to the aggregator with a vector $\pmb{y}$ in plaintext.
The TPA generates a functionally derived key that depends on $y$, denoted as $\texttt{dk}_{\pmb{y}}$.
The aggregator decrypts $ct_{\pmb{x}}$ using the received key denoted as $\texttt{dk}_{\pmb{y}}$.
As a result of the decryption, the aggregator obtains the result inner product of $\pmb{x}$ and $\pmb{y}$ in plaintext.
Notice that to securely apply FE cryptosystem, the TPA should not get access to encrypted $\pmb{x}$.

More formally in SIFE, the supported function is
$\langle\pmb{x},\pmb{y}\rangle = \sum^{\eta}_{i=1}(x_{i}y_{i})$, where $\pmb{x}$ and $\pmb{y}$ are two vectors of length $\eta$.
For a formal definition, we refer the reader to \cite{abdalla2015simple}. 
We briefly described the main algorithms as follows in terms of our system entities:

\noindent 1. $\mathcal{E}_\text{SIFE}.\text{Setup}$: Used by the TPA to generate a master private key and common public key pairs based on a given security parameter.

\noindent 2. $\mathcal{E}_\text{SIFE}.\text{DKGen}$: Used by the TPA. It takes the master private key and one vector $\pmb{y}$ as input, and generates a functionally derived key as output.

\noindent 3. $\mathcal{E}_\text{SIFE}.\text{Enc}$: Used by a party to output ciphertext of vector $\pmb{x}$ using the public key $\texttt{pk}^{\text{SIFE}}$.
We denote this as
$ct_{\pmb{x}}=\mathcal{E}_\text{SIFE}.\text{Enc}_{\texttt{pk}^{\text{SIFE}}}(\pmb{x})$

\noindent 4. $\mathcal{E}_\text{SIFE}.\text{Dec}$: Used by the aggregator. This algorithm takes the ciphertext, the public key and functional key for the vector $\pmb{y}$ as input, and returns the inner-product $\langle\pmb{x},\pmb{y}\rangle$.

\noindent\textbf{MIFE}($\mathcal{E}_\text{MIFE}$). 
We also make use of the $\mathcal{E}_\text{MIFE}$ cryptosystem,
which provides similar functionality to SIFE only that the private data $x$ comes from multiple parties.
The supported function is $\langle\{\pmb{x}_{i}\}_{i\in[n]},\pmb{y}\rangle = \sum_{i\in[n]}\sum_{j\in[\eta_{i}]}(x_{ij}y_{\sum^{i-1}_{k=1}\eta_{k}+j})
\;\text{s.t.}\;\; |\pmb{x}_i|=\eta_{i}, |\pmb{y}| = \sum_{i\in[n]}\eta_{i}, $
where $\pmb{x}_i$ and $\pmb{y}$ are vectors.
Accordingly, the MIFE scheme formally defined in \cite{abdalla2018multi} includes five algorithms briefly described as follows:

\noindent 1. $\mathcal{E}_\text{MIFE}.\text{Setup}$: Used by the TPA to generate a master private key and public parameters based on given security parameter and functional parameters such as the maximum number of input parties and the maximum input length vector of the corresponding parties.

\noindent 2. $\mathcal{E}_\text{MIFE}.\text{SKDist}$: Used by the TPA to deliver the secret key $\texttt{sk}^{\text{MIFE}}_{p_i}$ for a specified party $p_i$ given the master public/private keys. 

\noindent 3. $\mathcal{E}_\text{MIFE}.\text{DKGen}$: Used by the TPA. Takes the master public/private keys and vector $\pmb{y}$ as inputs, which is in plaintext and public, and generates a functionally derived key $\texttt{dk}_{\pmb{y}}$ as output. 

\noindent 4. $\mathcal{E}_\text{MIFE}.\text{Enc}$: Used by the aggregator to output ciphertext of vector $\textbf{x}_{i}$ using the corresponding secret key $\texttt{sk}^{\text{MIFE}}_{p_i}$. We denote this as $ct_{\pmb{x}_{i}}=\mathcal{E}_\text{MIFE}.\text{Enc}_{\texttt{sk}^{\text{MIFE}}_{p_i}}(\pmb{x}_{i})$.

\noindent 5. $\mathcal{E}_\text{MIFE}.\text{Dec}$: It takes the ciphertext set, the public parameters and functionally derived key $\texttt{dk}_{\pmb{y}}$ as input, and returns the inner-product $\langle\{\pmb{x}_{i}\}_{i\in[n]},\pmb{y}\rangle$.

We now introduce \textit{FedV} and explain how these cryptosystems are used to train multiple types of ML models.

\section{The Proposed \textit{FedV} Framework}
\label{sec:vfl}

\begin{figure*}
    \centering
    \includegraphics[scale=0.55]{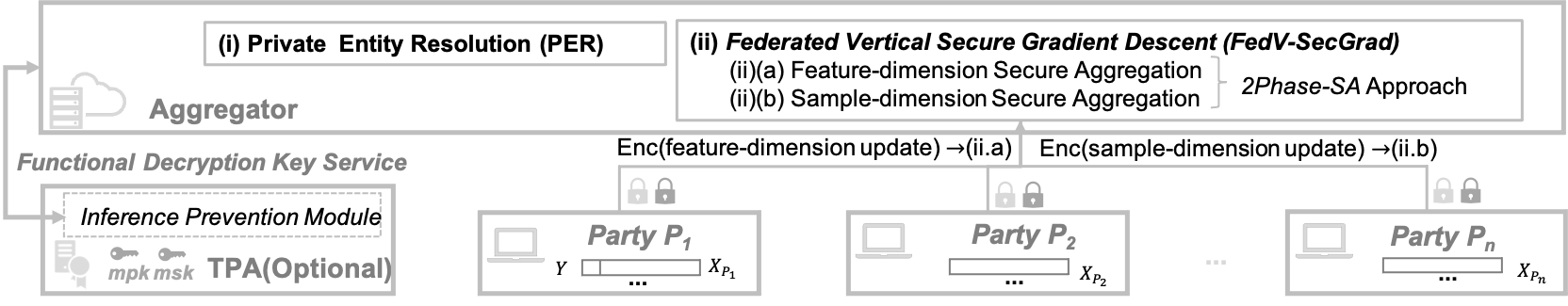}
    \vspace{-2mm}
    \caption{Overview of \textit{FedV} architecture: no peer-to-peer communication needed. We assume party $p_1$ owns the labels, while all other parties (i.e., $p_2, ..., p_n$) are passive parties. Note that crypto-infrastructure TPA component could be optional, which depends on the adopted FE schemes. In the TPA-free FE setting, the inference prevention module can be deployed at encryption entity, i.e., training parties in the FL.}
    
    \label{fig:framework}
    \vspace{-3mm}
\end{figure*}

We now introduce our proposed approach, \textit{FedV}, which is shown in \figurename\;\ref{fig:framework}.
\textit{FedV} has three types of entities: an \textit{aggregator}, a set of \textit{parties} and a \textit{third-party authority (TPA)} crypto-infrastructure to enable functional encryption.
The \textit{aggregator} orchestrates the private entity resolution procedure and coordinates the training process among the parties.
Each \textit{party} owns a training dataset which contains a subset of features and wants to collaboratively train a global model. 
We name \textit{parties} as follows: 
(i)~one \textit{active party} who has training samples with partial features and the class labels, represented as $p_1$ in Figure~\ref{fig:framework};
and
(ii)~multiple \textit{passive parties} who have training samples with only partial features.

\subsection{Threat Model and Assumptions}
\label{sec:vfl:threat}

The main goal of \textit{FedV} is to train an ML model protecting the privacy of the features provided by each party
without revealing beyond what is revealed by the model itself.
That is, \textit{FedV} enables \textit{privacy of the input}.
The goal of the adversary is to infer party's features.
We now present the assumptions for each entity in the system.

We assume an {honest-but-curious aggregator} who correctly follows the algorithms and protocols, but may try to learn private information from the aggregated model updates. The aggregator is often times run by large companies, where adversaries may
have a hard time modifying the protocol without been noticed by others.
    
With respect to the parties in the system, we assume a limited number of \textit{dishonest} parties who 
may try to infer the honest parties' private information.
Dishonest parties may collude with each other to try to obtain features from other participants.
In \textit{FedV}, the number of such parties is bounded by $m-1$ out of $m$ parties.
We also assume that the aggregator and parties do not collude. 

To enable functional encryption, a TPA may be used.
At the time of completion of this work, new and promising cryptosystems that remove the TPA have been proposed \cite{chotard2018decentralized, abdalla2019decentralizing}.
These cryptosystems do not require a trusted TPA.
If a cryptosystem that uses a TPA is used, this entity needs to be fully trusted by other entities in the system
to provide functional derived keys uniquely to the aggregator.
In real-world scenarios, different sectors already have entities that can take the role of a TPA.
For example, central banks of the banking industry often play a role of a fully trusted entity.
In other sectors third-party companies such as consultant firms can run the TPA.



We assume that secure channels are in place; hence, \textit{man-in-the-middle} and \textit{snooping} attacks are not feasible.
Finally, denial of service attacks and backdoor attacks  \cite{chen2018detecting, bagdasaryan2018backdoor} where parties try to cause the final model to create a targeted misclassification are outside the scope of this paper.

\subsection{Overview of FedV}


\textit{FedV} enables VFL without a need for any peer-to-peer communication resulting in a drastic reduction in training time and amounts of data that need to be transferred.
We first overview the entities in the system and explain how they interact under our proposed two-phase secure aggregation technique that makes these results possible.

Algorithm \ref{alg:vfl} shows the operations followed by \textit{FedV}.
First crypto keys are obtained by all entities in the system. 
After that, to align the samples of each parties, a private entity resolution process as defined in \cite{schnell2011novel,hardy2017private} (see section \ref{sec:bg:vfl}) takes place.
Here, each party receives an entity resolution vector, $\pmb{\pi}_i$, and shuffles its local data samples under the aggregator's orchestration.
This results in parties having all records appropriately aligned before the training phase starts.



\let\oldnl\nl
\newcommand{\nonl}{\renewcommand{\nl}{\let\nl\oldnl}}
\begin{algorithm}[!t]
    \SetAlgoLined
    \caption{\textit{FedV} Framework \newline
    {\small 
    {\bf Inputs:} $s$ := batch size, $maxEpochs$, and $S$ := total batches per epoch, $d$ := total number features. \newline
    {\bf System Setup:} 
    TPA initializes cryptosystems, delivers public keys and a secret random seed $r$ to each party.
    }
    }
    \label{alg:vfl}
    \small
    \SetKwProg{Aggregator}{Aggregator}{}{}
    \SetKwProg{Party}{Party}{}{}
    
    \nonl\Party{}{
        Re-shuffle its samples using the received entity resolution vector $(\pmb{\pi}_1, ..., \pmb{\pi}_n)$ \;\label{alg:vfl:per}
        Use $r$ to generate it's one-time password chain
    }
    
    \nonl\Aggregator{}{
        $\pmb{w} \gets$ random initialization\;
        \ForEach{epoch in $maxEpochs$}{
            $\nabla E(\pmb{w}) \leftarrow $ FedV-SecGrad($epoch, s, S, d, \pmb{w}$)\; \label{alg:vfl:ln:secgrad}
                $\pmb{w} \leftarrow \pmb{w} - \alpha\nabla E(\pmb{w})$; \label{alg:vfl:ln:descent}
        }
        \Return $\pmb{w}$
    }
\end{algorithm}

%



The training process by
executing the \textit{Federated Vertical Secure Gradient Descent} (\textit{FedV-SecGrad}) procedure, which is the \textit{core novelty} of this paper.
\textit{FedV-SecGrad} is called at the start of each epoch to securely compute the gradient of the loss function $E$ based on $\mathcal{D}$.  
\textit{FedV-SecGrad} consists of a two-phased secure aggregation operation that enables the computation of gradients and requires the parties to perform a sample-dimension and feature-dimension encryption (see Section \ref{sec:vfl:detail}).
The resulting cyphertexts are then sent to the aggregator.

Then, the aggregator generates an \textit{aggregation vector} to compute the inner products and sends it to the TPA.
For example upon receiving two ciphertexts $ct_1$, $ct_2$, the aggregator generates an aggregation vector $(1,1)$ and sends it to the TPA,
which returns the functional key to the aggregator to compute the inner product between $(ct_1,ct_2)$ and $(1,1)$.
Notice that the TPA doesn't get access to $ct_1$, $ct_2$ and the final result of the aggregation.
Note that the aggregation vectors \textit{do not} contain any private information;
they only include the weights used to aggregate ciphertext coming from parties.

To prevent inference threats that will be explained in detail in Section~\ref{sec:vfl:dynamic},
once the TPA gets an aggregation vector, it is inspected by its \textit{Inference Prevention Module} (IPM),
which is responsible for making sure the vectors are adequate.
If the IPM concludes that the aggregation vectors are valid, the TPA provides the functional decryption key to the aggregator.
Notice that \textit{FedV} is compatible with TPA-free FE schemes \cite{abdalla2019decentralizing,chotard2018decentralized}, where parties collaboratively generate the functional decryption key. In that scenario, the IPM can also be deployed at each training party.
Using decryption key, the aggregator then obtains the result of the corresponding inner product via decryption.
As a result of these computations, the aggregator can obtain the exact gradients that can be used for any gradient-based step to update the ML model.
Line~\ref{alg:vfl:ln:descent} of Algorithm~\ref{alg:vfl} uses stochastic gradient descent (SGD) method to illustrate the update step of the ML model. 
%
%
We present \textit{FedV-SecGrad} in details in Section~\ref{sec:vfl:detail}.

\section{Vertical Training Process: \textit{FedV-SecGrad}}
\label{sec:vfl:detail}

We now present in detail our \textit{federated vertical secure gradient descent (FedV-SecGrad)} and its supported ML models, as captured in the following claim.


\newtheorem{claim}{Claim}
\begin{claim}
    \label{theo:gp_vfl_theorem}
    \textit{
    FedV-SecGrad is a generic approach to securely compute gradients of an ML objective with a
    prediction function that can be written as $f(\pmb{x};\pmb{w}):=g(\pmb{w}^{\intercal}\pmb{x})$, where $g: \mathbb{R} \rightarrow \mathbb{R}$ is a differentiable function, $\pmb{x}$ and $\pmb{w}$ denote the feature vector and the model weights vector, respectively.
    }
\end{claim}

ML objective defined in Claim~\ref{theo:gp_vfl_theorem} covers many classical ML models including nonlinear models, such as logistic regression, SVMs, etc.
When $g$ is the identity function, the ML objective $f$ reduces to a linear model, which will be discussed in Section~\ref{sec:linear}.
When $g$ is not the identity function, Claim~\ref{theo:gp_vfl_theorem} covers a special class of nonlinear ML model; for example, when $g$ is the sigmoid function, our defined ML objective is a logistic classification/regression model.
We demonstrate how \textit{FedV-SecGrad} is extended to nonlinear models in Section~\ref{sec:example-algorithms}.
Note that in Claim~\ref{theo:gp_vfl_theorem}, we deliberately omit the regularizer $R$ commonly used in an ML (see equation \eqref{equ:train_loss}), because common regularizers only depend on model weights $\pmb{w}$; it can be computed by the aggregator independently.
We provide details of how logistic regression models are covered by Claim~\ref{theo:gp_vfl_theorem} in Appendix~\ref{appendix:proof}

\subsection{\textit{FedV-SecGrad} for Linear Models}\label{sec:linear}

We first present \textit{FedV-SecGrad} for linear models, where $g$ is the identity function and the loss is the mean-squared loss.
The target loss function then becomes $E(\pmb{w}) = \tfrac{1}{2n}\tsum^{n}_{i=1}(y^{(i)}- \pmb{w}^{\intercal}\pmb{x}^{(i)})^2.$
We observe that the gradient computations over vertically partitioned data, $\nabla E(\pmb{w})$, can be reduced to two types of operations: 
(\romannumeral1) \textit{feature-dimension aggregation}
and 
(\romannumeral2) \textit{sample/batch-dimension aggregation}.
To perform these two operations, \textit{FedV-SecGrad} follows a
\textit{two-phased secure aggregation (2Phased-SA)} process.
Specifically, the \textit{feature dimension SA} securely aggregates several batches of training data that belong to different parties in feature-dimension to acquire the value of $y^{(i)}- \pmb{x}^{(i)}\pmb{w}$ for each data sample as illustrated in \eqref{eq:linear_g_update}, while the 
\textit{sample dimension SA} can securely aggregate one batch of training data owned by one party in sample-dimension with the weight of $y^{(i)}- \pmb{x}^{(i)}\pmb{w}$ for each sample, to obtain the batch gradient $\nabla E_{\mathcal{B}}(\pmb{w})$.
The communication between the \textit{parties} and the \textit{aggregator} is a one-way interaction requiring a single message.


We use a simple case of two parties to illustrate the proposed protocols, 
where $p_1$ is the active party and $p_2$ is a passive party.
Recall that the training batch size is $s$ and the total number of features is $d$. 
Then the current training batch samples for $p_1$ and $p_2$ can be denoted as $\mathcal{B}^{s\times m}_{p_1}$ and $\mathcal{B}^{s\times (d-m)}_{p_2}$ as follows:
\begin{align*}
    \mathcal{B}^{s\times m}_{p_1}\quad\qquad&\qquad\qquad \mathcal{B}^{s\times (d-m)}_{p_2} \\
    \begin{bmatrix}
        y^{(1)}       \\
        \vdots   \\
        y^{(s)}       
    \end{bmatrix}
    \begin{bmatrix}
        x^{(1)}_1  & \dots & x^{(1)}_m      \\
        \vdots  &  \ddots & \vdots  \\
        x^{(s)}_1  &  \dots & x^{(s)}_m      
    \end{bmatrix}
    & 
    \begin{bmatrix}
        x^{(1)}_{m+1}  & \dots & x^{(1)}_{d}      \\
        \vdots  &  \ddots & \vdots  \\
        x^{(s)}_{m+1}  &  \dots & x^{(s)}_{d}      
    \end{bmatrix} 
\end{align*}

%

\noindent\textit{Feature dimension SA}.
The goal of \textit{feature dimension SA} is to securely aggregate the sum of a group of `partial models' $\pmb{x}^{(i)}_{p_i}\pmb{w}_{p_i}$, from multiple parties without disclosing the inputs to the \textit{aggregator}.
Taking the $s^{\text{th}}$ data sample in the batch as an example,
the aggregator is able to securely aggregate
$\sum_{k=1}^mw_kx^{(s)}_k -y^{(s)}$+$\sum_{k=m+1}^d w_kx^{(s)}_k$.
For this purpose, the active party and all other passive parties perform slightly different pre-processing steps before invoking \textit{FedV-SecGrad}.
The active party, $p_1$, appends a vector with labels $y$ to obtain $\pmb{x}^{(i)}_{p_1}\pmb{w}_{p_1} - y^{(i)}$ as its `partial model'.
For the passive party $p_2$, its `partial model' is defined by $\pmb{x}^{(i)}_{p_2}\pmb{w}_{p_2}$. 
Each party $p_i$ encrypts its `partial model' using the MIFE encryption algorithm with its public key $pk^{\text{MIFE}}_{p_i}$, and sends it to the aggregator.

Once the aggregator receives the partial models, it prepares a fusion vector $\pmb{v}_{\mathcal{P}}$ of size equal to the number of parties to perform the aggregation and sends it to the TPA to request a function key  $\text{sk}^{\text{MIFE}}_{\pmb{v}_{\mathcal{P}}}$.
With the received key $\text{sk}^{\text{MIFE}}_{\pmb{v}_{\mathcal{P}}}$,
the aggregator can obtain
the aggregated sum of the elements of $\pmb{w}^{m\times 1}_{p_1}\mathcal{B}^{s\times m}_{p_1} - \pmb{y}^{1\times s}$ and $\pmb{w}^{(d-m)\times 1}_{p_2}\mathcal{B}^{s\times (d-m)}_{p_2}$ in the feature dimension.

It is easy to extend the above protocol to a general case with $n$ parties. 
In this case, the fusion vector $\pmb{v}$
can be set as a binary vector with $n$ elements, where one indicates that the aggregator has received the replies from the corresponding party, and zero indicates otherwise. 
In this case, the aggregator gives equal fusion weights to all the replies for the feature dimension aggregation.
We discuss the case where only a subset of parties replies in detail in Section \ref{sec:vfl:dynamic}.

\noindent\textit{Sample dimension SA}.
The goal of the \textit{sample dimension SA} is to securely aggregate the batch gradient. 
For example, considering the first feature weight $w_1$ for data sample owned by $p_1$, the aggregator is able to securely aggregate $\nabla E(w_1)=\sum_{k=1}^sx^{(k)}_1u_{k}$ via \textit{sample dimension SA} where $\pmb{u}$ is the aggregation result of \textit{feature dimension SA} discussed above. 
This SA protocol requires the party to encrypt its batch samples using the SIFE cryptosystem with its public key $pk^{\text{SIFE}}$.
Then, the \textit{aggregator} exploits the results of the \textit{feature dimension SA}, i.e., an element-related weight vector $\pmb{u}$
to request a function key $\text{sk}^{\text{SIFE}}_{\pmb{u}}$ from the TPA.
With the function key $\text{sk}^{\text{SIFE}}_{\pmb{u}}$, the aggregator is able to decrypt the ciphertext
and acquire the batch gradient $\nabla E(\pmb{w})$.



\noindent\textbf{Detailed Execution of the FedV-SecGrad Process.}
As shown in Algorithm~\ref{alg:vfl}, the general \textit{FedV}  adopts a mini-batch based SGD algorithm to train a ML model in a VFL setting. After system setup, all parties use the random seed provided by the TPA to generate a one-time-password sequence \cite{haller1998one} that will be used to generate batches during the training process.
Then, the training process can begin.

At each training epoch, the \textit{FedV-SecGrad} approach specified in Procedure~\ref{alg:sg} is invoked in line \ref{alg:vfl:ln:secgrad} of Algorithm~\ref{alg:vfl}.
The aggregator queries the parties with the current model weights, $\pmb{w}_{p_i}$.
To reduce data transfer and protect against inference attacks\footnote{In this type of attack, a party may try to find out if its features are more important than those of other parties. This can be easily inferred in linear models.}, the aggregator only sends each party the weights that pertain to its partial feature set. We denote these partial model weights as 
$\pmb{w}_{p_i}$ in line \ref{alg:sg:ln:weight}.

In Algorithm \ref{alg:vfl}, each party uses a random seed $r$ to generate its one-time password chain.
For each training epoch, each party uses the one-time-password chain associated with the training epoch to randomly select the samples that are going to be included in a batch for the given batch index, as shown in line \ref{alg:sg:ln:getbatch}.
In this way, the aggregator never gets to know what samples are included in a batch, thus preventing inference attacks (see Section \ref{sec:sp}). 

Then each party follows the feature-dimension and sample-dimension encryption process shown in lines \ref{alg:sg:ln:fsa:active}, \ref{alg:sg:ln:fsa} and \ref{alg:sg:ln:ssa} of Procedure~\ref{alg:sg}, respectively.
As a result, each party's local `partial model' is encrypted and the two ciphertexts, $\pmb{c}_{\text{fd}}$ and $ \pmb{c}_{\text{sd}}$, are sent back to the aggregator.
The aggregator waits for a pre-defined duration for parties' replies, denoted as two sets of corresponding ciphertexts $\mathcal{C}^{\text{fd}}$ and $ \mathcal{C}^{\text{sd}}$.
Once this duration has elapsed, it continues the training process by performing the following secure aggregation steps.
First, the feature dimension SA, is performed.
For this purpose, in line \ref{alg:sg:ln:init1}, vector $\pmb{v}$ is initialized with all-one vector and is updated to zeros for not responding parties, as in line \ref{alg:sg:ln:v}.
This vector provides the weights for the inputs of the received encrypted `partial models'.
Vector $\pmb{v}$ is sent to the TPA 
that verifies the suitability of the vector (see Section \ref{sec:vfl:dynamic}).
If $\pmb{v}$ is suitable, the TPA returns the private key $\text{dk}_{\pmb{v}}$ to perform the decryption.
The feature dimension SA, is completed in line \ref{alg:sg:ln:sa-f}, where the MIFE based decryption takes place resulting in 
$\pmb{u}$ that contains the aggregated weighted feature values of $s$-th batch samples.
Then the sample dimension, SA, takes place, where
the aggregator uses $\pmb{u}$ as an aggregation vector and sends it to the TPA to obtain a functional key $\text{dk}_{ \pmb{u}}$.
The TPA verifies the validity of $\pmb{u}$ and returns the key if appropriate (see Section \ref{sec:vfl:dynamic}).
Finally, the aggregated gradient update $\nabla E_{b_\text{idx}}(\pmb{w})$ is computed as in lines \ref{alg:sg:ln:sa-s} and \ref{alg:sg:ln:gradient} by performing a SIFE decryption using $\text{dk}_{\pmb{u}}$.

\SetAlgorithmName{Procedure}{procedure}{}
\definecolor{arsenic}{rgb}{0.23, 0.27, 0.29}
\newcommand\mycommfont[1]{\scriptsize\ttfamily\textcolor{arsenic}{#1}}
\SetCommentSty{mycommfont}
\begin{algorithm}[!t]
    \small
    \caption{FedV-SecGrad
    {
    }} 
    \label{alg:sg}
    \SetKwRepeat{Do}{do}{while}
    \SetKwProg{Fn}{function}{}{}
    \SetKwProg{Aggregator}{Aggregator}{}{}
    \SetKwProg{Party}{Party}{}{}
    \SetKwProg{TPA}{TPA}{}{}
    \nonl\Aggregator{FedV-SecGrad($epoch, s, S, d, \pmb{w}$)}{ 
        generate batch indices $\{1, ..., m\}$ according to $S$\;
        divide model $\pmb{w}$ into partial model $\pmb{w}_{p_i}$ for each $p_i$\;\label{alg:sg:ln:weight}
        \ForEach{$b_{\text{idx}} \in \{1, ..., m\}$}{
            \tcc{initialize inner product vectors}
            $\pmb{v}$=$\pmb{1}^{n}$ \tcp*[l]{feature dim. aggregation vector} \label{alg:sg:ln:init1}
            $\pmb{u}$=$\pmb{0}^{s}$ \tcp*[l]{sample dim. aggregation vector} \label{alg:sg:ln:init2}
            \tcc{initialize matrices for replies}
            $\mathcal{C}^{\text{fd}}$=$\pmb{0}^{n\times s} $ \tcp*[l]{ciphertexts of feature dim.} \label{alg:sg:ln:init3}
            $\mathcal{C}^{\text{sd}}$=$\pmb{0}^{s\times d}$ \tcp*[l]{ciphertexts of sample dim.} \label{alg:sg:ln:init4}
        \ForEach{$p_i \in \mathcal{P}$}{
            $\mathcal{C}^{\text{fd}}_{i,\cdot}, \mathcal{C}^{\text{sd}}_{\cdot, j} \gets \textit{query-party}(\pmb{w}_{p_i}, b_{\text{idx}}, s)$\;
            \lIf{$p_i$ did not reply}{$v_i=0$ \label{alg:sg:ln:v}}
        }
        $\texttt{dk}^{\text{MIFE}}_{\pmb{v}} \gets \textit{query-key-service}(\pmb{v}, \mathcal{E}_{\text{MIFE}})$ \; \label{alg:sg:ln:key-v}
        \ForEach{$k \in \{1,...,s\}$}{
                
            $u_k \leftarrow \mathcal{E}_\text{MIFE}.\text{Dec}_{\texttt{dk}^{\text{MIFE}}_{\pmb{v}}}(\{\mathcal{C}^{\text{fd}}_{i,k}\}_{i\in\{1,...,n\}})$ \label{alg:sg:ln:sa-f}
        }
        $\texttt{dk}^{\text{SIFE}}_{\pmb{u}} \gets \textit{query-key-service}(\pmb{u}, \mathcal{E}_{\text{SIFE}})$\;\label{alg:sg:ln:key-u}
        \ForEach{$j \in \{1, ..., d\}$}{
            $\nabla E^{'}(\pmb{w})_{j} \leftarrow \mathcal{E}_\text{SIFE}.\text{Dec}_{\texttt{dk}^{\text{SIFE}}_{\pmb{u}}}(\{\mathcal{C}^{\text{sd}}_{k,j}\}_{k\in\{1,...,s\}})$\label{alg:sg:ln:sa-s}
        }
        $\nabla E_{b_{\text{idx}}}(\pmb{w}) \leftarrow \nabla E^{'}(\pmb{w}) + \lambda\nabla R(\pmb{w})$;
        }\label{alg:sg:ln:gradient}
        \Return $\frac{1}{m}\sum_{b_{\text{idx}}}\nabla E_{b_{\text{idx}}}(\pmb{w})$
    }
    \nonl\Party{}{
        \nonl\textit{\textbf{Inputs:}} $\mathcal{D}_{p_i}$:=pre-shuffled party's dataset, $\texttt{sk}^{\text{MIFE}}_{p_i},\texttt{pk}^{\text{SIFE}}$:=public keys of party\;
        \Fn{query-party($\pmb{w}_{p_i}, b_{\text{idx}}, s$)}{
            \tcp{get batch using one-time-password chain}
            $\mathcal{B}_{p_i} \leftarrow
            get\_batch(b_{\text{idx}}, s,  \mathcal{D}_{p_i})$\;\label{alg:sg:ln:getbatch}
            \If{$p_i$ is active party}{
                $\pmb{ct}_{\text{fd}} \leftarrow \mathcal{E}_\text{MIFE}.\text{Enc}_{\texttt{sk}^{\text{MIFE}}_{p_i}}(\pmb{w}_{p_i}\mathcal{B}_{p_i}-\pmb{y})$; \label{alg:sg:ln:fsa:active}
            }
            \lElse {
                $\pmb{ct}_{\text{fd}} \leftarrow \mathcal{E}_\text{MIFE}.\text{Enc}_{\texttt{sk}^{\text{MIFE}}_{p_i}}(\pmb{w}_{p_i}\mathcal{B}_{p_i})$\label{alg:sg:ln:fsa}
            }
            $\pmb{ct}_{\text{sd}} \leftarrow \mathcal{E}_\text{SIFE}.\text{Enc}_{\texttt{pk}^{\text{SIFE}}}(\mathcal{B}_{p_i})$ \tcp*[l]{in sample dim.}\label{alg:sg:ln:ssa}
            \Return $(\pmb{ct}_{\text{fd}}, \pmb{ct}_{\text{sd}})$ to the aggregator\;
        }
    }
    \nonl\TPA{}{
        \nonl \textit{\textbf{Inputs:}} $n$:=number of parties, $t$:=min threshold of parties, $s$:=bath size\;
        \Fn{query-key-service($\pmb{v}|\pmb{u}, \mathcal{E}$)}{
            \lIf{IPM($\pmb{v}|\pmb{u},\mathcal{E}$)}{
                \Return $\mathcal{E}.DKGen(\pmb{v}|\pmb{u})$
            }
            \lElse{\Return `exploited vector'}
        }
        \Fn{IPM($\pmb{v}|\pmb{u}, \mathcal{E}$)}{
            \If{$\mathcal{E}$ is $\mathcal{E}_{\text{MIFE}}$}{
                \lIf{$|\pmb{v}|=n$\textbf{ and }$sum(\pmb{v})>t$}{\Return \textbf{true}}\label{alg:sg:ln:ipm-v}
                \lElse{\Return \textbf{false}}
            }
            \ElseIf{$\mathcal{E}$ is $\mathcal{E}_{\text{SIFE}}$}{
                \lIf{$|\pmb{u}|=s$}{\Return \textbf{true}}\label{alg:sg:ln:ipm-u}
                \lElse{\Return \textbf{false}}
            }
        }
    }
\end{algorithm}


\subsection{\textit{FedV} for Non-linear Models} \label{sec:example-algorithms}

In this section, we extend \textit{FedV-SecGrad} to compute gradients of non-linear models, i.e., when $g$ is not the identity function in Claim~\ref{theo:gp_vfl_theorem}, without the help of Taylor approximation.
For non-linear models, \textit{FedV-SecGrad} requires the active party to share labels with the aggregator in plaintext.   
Since $g$ is not the identity function and may be nonlinear,
the corresponding gradient computation does not consist only linear operations.
We present the differences between Procedure~\ref{alg:sg} and \textit{FedV-SecGrad} for non-linear models in Procedure~\ref{alg:sg:non-liner}.
Here, we briefly analyze the extension on logistic models and SVM models.
More details can be found in Appendix~\ref{appendix:proof}.

\LinesNotNumbered
\begin{algorithm}[!t]
    \small
    \caption{FedV-SecGrad for Non-linear Models. \newline
    {\small {\bf Note:} For conciseness, operations shared with Procedure~\ref{alg:sg} are not presented. Please refer to that Procedure }
    }
    \label{alg:sg:non-liner}
    \SetKwProg{Fn}{function}{}{}
    \SetKwProg{Aggregator}{Aggregator}{}{}
    \SetKwProg{Party}{Party}{}{}
    \Aggregator{}{ 
        \setcounter{AlgoLine}{11}
        \nl\ForEach{$k \in \{1,...,s\}$}{
                
            \nl $z_k \leftarrow \mathcal{E}_\text{MIFE}.\text{Dec}_{\texttt{dk}^{\text{MIFE}}_{\pmb{v}}}(\{\mathcal{C}^{\text{fd}}_{i,k}\}_{i\in\{1,...,n\}})$
        }
        \nl $\pmb{u} \leftarrow g(\pmb{z}) - \pmb{y}$\tcp*[l]{adapted to a specific loss}\label{alg:sg:ln:u}
    }
    \Party{}{ 
        \setcounter{AlgoLine}{19}
        \nl $\mathcal{B}_{p_i} \leftarrow
            get\_batch(b_{\text{idx}}, s,  \mathcal{D}_{p_i})$\;
        \nl $\pmb{ct}_{\text{fd}} \leftarrow \mathcal{E}_\text{MIFE}.\text{Enc}_{\texttt{sk}^{\text{MIFE}}_{p_i}}(\pmb{w}_{p_i}\mathcal{B}_{p_i})$\;
        \nl $\pmb{ct}_{\text{sd}} \leftarrow \mathcal{E}_\text{SIFE}.\text{Enc}_{\texttt{pk}^{\text{SIFE}}}(\mathcal{B}_{p_i})$ \tcp*[l]{in sample dimension}
        \nl \lIf{$p_i$ is active party}{\Return $(\pmb{ct}_{\text{fd}}, \pmb{ct}_{\text{sd}}, \pmb{y})$ to the aggregator}
        \nl \lElse{\Return $(\pmb{ct}_{\text{fd}}, \pmb{ct}_{\text{sd}})$ to the aggregator}
    }
\end{algorithm}

\noindent \textit{Logistic Models}.
We now rewrite the prediction function $f(\pmb{x};\pmb{w}) = \frac{1}{1+e^{-\pmb{w}^{\intercal}\pmb{x}}}$ as $g(\pmb{w}^{\intercal}\pmb{x})$, where $g(\cdot)$ is the sigmoid function, i.e.,  $g(z)=\frac{1}{1+e^{-z}}$. 
If we consider classification problem and hence use cross-encropy loss, the gradient computation over a mini-batch $\mathcal{B}$ of size $s$ can be described as
$\nabla E_{\mathcal{B}}(\pmb{w}) = \frac{1}{s}\sum_{i \in \mathcal{B}}(g(\pmb{w}^{(i)\intercal}\pmb{x}^{(i)})) - y^{(i)})\pmb{x}^{(i)}$.
The \textit{aggregator} is able to acquire $z^{(i)}=\pmb{w}^{(i)\intercal}\pmb{x}^{(i)}$ following the \textit{feature dimension SA} process. 
With the provided labels, it can then compute $u_{i} = g(\pmb{z}) - y^{(i)}$ as in line~\ref{alg:sg:ln:u} of Procedure~\ref{alg:sg:non-liner}.
Note that line~\ref{alg:sg:ln:u} is specific for the adopted cross-entropy loss function.  
If another loss function is used, we need to update line~\ref{alg:sg:ln:u} accordingly.
Finally, \textit{sample dimension SA} is applied to compute $\nabla E_{\mathcal{B}}(\pmb{w})=\sum_{i \in \mathcal{B}}u_{i}\pmb{x}^{(i)}$.
\textit{FedV-SecGrad} also provides an alternative approach for the case of restricting label sharing, where the logistic computation is transferred to linear computation via Taylor approximation, as used in existing VFL solutions \cite{hardy2017private}.
Detailed specifications of the above approaches are provided in Appendix \ref{appendix:proof}.

\noindent \textit{SVMs with Kernels}. 
SVM with kernel is usually used when data is not linearly separable. 
We first discuss linear SVM model.
When it uses squared hinge loss function and its objective is to minimize 
$\frac{1}{n}\sum_{i\in\mathcal{B}}\left(\max(0, 1-y^{(i)}\pmb{w}^{(i)\intercal}\pmb{x}^{(i)})\right)^2$.
The gradient computation over a mini-batch $\mathcal{B}$ of size $s$ can be described as
$\nabla E_{\mathcal{B}}(\pmb{w}) = \frac{1}{s}\sum_{i \in \mathcal{B}}-2y^{(i)}(\max(0, 1-y^{(i)}\pmb{w}^{(i)\intercal}\pmb{x}^{(i)}))\pmb{x}^{(i)}$.
With the provided labels and acquired $\pmb{w}^{(i)\intercal}\pmb{x}^{(i)}$,  Line~\ref{alg:sg:ln:u} of Procedure~\ref{alg:sg:non-liner} can be updated so that the aggregator computes $u_i= -2y^{(i)}\max(0, 1-y^{(i)}\pmb{w}^{(i)\intercal}\pmb{x}^{(i)})$ instead.
Now let us consider the case where SVM uses nonlinear kernels. 
Suppose the prediction function is $f(\pmb{x};\pmb{w}) = \sum^{n}_{i=1} w_iy_ik(\pmb{x}_i, \pmb{x})$, where $k(\cdot)$ denotes the corresponding kernel function. 
As nonlinear kernel functions, such as polynomial kernel $(\pmb{x}^{\intercal}_i\pmb{x}_j)^d$, sigmoid kernel $tanh(\beta \pmb{x}^{\intercal}_i\pmb{x}_j + \theta)$ ($\beta$ and $\theta$ are kernel coefficients), are based on inner-product computation which is supported by our \textit{feature dimension SA} and \textit{sample dimension SA} protocols, these kernel matrices can be computed before the training process begins. 
And the aforementioned objective for SVM with nonlinear kernels will be reduced to SVM with linear kernel case with the pre-computed kernel matrix. Then the gradient computation process for these SVM models will be reduced to a gradient computation of a standard linear SVM,
which can clearly be supported by \textit{FedV-SecGrad}.

\subsection{Enabling Dynamic Participation in \textit{FedV} and Inference Prevention}
\label{sec:vfl:dynamic}

In some applications, parties may have glitches in their connectivity that momentarily inhibit their communication with the aggregator.
The ability to easily recover from such disruptions, ideally without losing the computations from all other parties, would help reduce the training time. \textit{FedV} allows a limited number of non-active parties to dynamically drop out and re-join during the training phase.
This is possible because \textit{FedV} requires neither sequential peer-to-peer communication among parties
nor re-keying operations when a party drops.
To overcome missing replies,
\textit{FedV} allows the aggregator to set the corresponding element in $\pmb{v}$ as zero (Procedure \ref{alg:sg}, line \ref{alg:sg:ln:v}).



\noindent\textbf{Inference Threats and Prevention Mechanisms.}
The dynamic nature of the inner product aggregation vector in Procedure \ref{alg:sg}, line \ref{alg:sg:ln:v},
may enable the inference attacks below, where the aggregator is able to isolate the inputs from a particular party.
We analyze two potential inference threats and show how \textit{FedV} design is resilient against them.

An honest-but-curious aggregator may be able to analyze the traces where some parties drop off; in this case, the resulting aggregated results will uniquely include a subset of replies making it easier to infer the input of a party.
This attack is defined as follows:

\begin{definition}[Inference Attack]
\label{def:infer_attack}
An inference attack carried by an adversary to infer party $p_i$ input $\pmb{w}_{p_{i}}^{\intercal}\pmb{x}^{(i)}_{p_{i}}$ or party's local features $\pmb{x}^{(i)}_{p_{i}}$ without directly accessing them.
\end{definition}
Here, we briefly analyze this threat from the feature and sample dimensions separately, and show how to prevent this type of attack even under the case of an actively curious aggregator. We formally prove the privacy guarantee of \textit{FedV} in Section~\ref{sec:sp}.

\noindent\textit{Feature dimension aggregation inference:}
To better understand this threat, let's consider an active attack where a curious aggregator obtains a function key, $\text{dk}_{\pmb{v}_{\text{exploited}}}$ by a manipulated vector such as $\pmb{v}_{\text{exploited}}=(0, ..., 0, 1)$ to infer the last party's input that corresponds to a target vector $\pmb{w}_{p_{n}}^{\intercal}\pmb{x}^{(i)}_{p_{n}}$ because the inner-product $\pmb{u}=\langle \pmb{w}^{\intercal}_{p_i}\pmb{x}^{(i)}_{p_i},\pmb{v}_{\text{exploited}}\rangle$ is known to the aggregator.

\noindent\textit{Sample dimension aggregation inference:}
An actively curious aggregator may decide to isolate a single sample by 
requesting a key that has fewer samples.
In particular, rather than requesting a key for $\pmb{u}$ of size $s$ (Procedure \ref{alg:sg} line \ref{alg:sg:ln:key-u}),
the curious aggregator may select a subset of $s$ samples, and in the worst case, a single sample. After the aggregation of this subset of samples, the aggregator may infer one feature value of a target data sample.

To mitigate the previous threats, the Inference Prevention Module (IPM) takes two parameters: $t$, a scalar that represents the minimum number of parties for which the aggregation is required,
and
$s$, which is the number of batch samples to be included in a sample aggregation.
For a feature aggregation, the IPM verifies that the vector's size is $n=\vert v \vert$, to ensure it is well formed according to Procedure \ref{alg:sg}, line \ref{alg:sg:ln:ipm-v}. Additionally, it verifies that the sum of its elements is greater than or equal to $t$ to ensure that at least the minimum tolerable number of parties' replies are aggregated. If these conditions hold, the TPA can return the associated functional key to the aggregator. 
Finally, to prevent sample based inference threats, 
the aggregator needs to verify that vector $\pmb{u}$ in Procedure \ref{alg:sg}, line \ref{alg:sg:ln:ipm-u} needs to always be equal to the predefined batch size $s$. 
By following this procedure the IPM ensures that the described active and passive inference attacks are thwarted so as to ensure the data of each party is kept private throughout the training phase. 

Another potential attack to infer the same target sample $\pmb{x}^{(\text{target})}$ is to utilize two manipulated vectors in subsequent training batch iterations, for example,
 $\pmb{v}^{\text{batch $i$}}_{\text{exploited}}=(1, ..., 1, 1)$ and $\pmb{v}^{\text{batch $i$+$1$}}_{\text{exploited}}=(1, ..., 1, 0)$ in training batch iteration $i$ and $i$+$1$, respectively.
Given results of $\langle \pmb{w}\pmb{x}^{(\text{target})},\pmb{v}^{\text{batch $i$}}_{\text{exploited}}\rangle$ and $\langle \pmb{w}\pmb{x}^{(\text{target})},\pmb{v}^{\text{batch $i$+$1$}}_{\text{exploited}}\rangle$, in theory the curious aggregator could subtract the latter one from the first to infer the target sample.
The IPM cannot prevent this attack, hence, we incorporate a random-batch selection process to address it.

\textit{FedV} incorporates a random-batch selection process that
makes it resilient against this threat. In particular, we incorporate randomness in the process of selecting data samples ensuring that the aggregator does not know if one sample is part of a batch or not.
Samples in each mini-batch are selected by parties according to a one-time password.
Due to this randomness, data samples included in each batch can be different.
Even if a curious aggregator computes the difference between two batches as described above, it cannot tell if the result corresponds to the same data sample or not, and no inference can be performed.
As long as the aggregator does not know the one-time password chain used to generate batches, the aforementioned attack is not possible. 
In summary, it is important for the one-time password to be kept secret by all parties from the aggregator.

\section{Security and Privacy Analysis}

\label{sec:sp}
Recall that the goal of \textit{FedV} is to train an ML model protecting the privacy of the features provided by each party without revealing beyond what is revealed by the model itself. In other words, \textit{FedV} protects the privacy of the input.
In this section, we formally prove the security and privacy guarantees of \textit{FedV} with respect to this goal.
First, we introduce the following lemmas with respects to the security of party input in the secure aggregation, security and randomness of one-time password (OTP) based seed generation, and solution of the non-homogeneous system to assist the proof of privacy guarantee of \textit{FedV} as shown in \textsc{Theorem}~\ref{theorem:privacy}.

\begin{lemma}[Security of Party Input]
\label{lemma:security}
The encrypted party's input in the secure aggregation of \textit{FedV} has ciphertext indistinguishability and is secure against adaptive corruptions under the classical DDH assumption.
\end{lemma}
The formal proof of \textsc{Lemma}~\ref{lemma:security} is presented in the functional encryption schemes \cite{abdalla2015simple, abdalla2018multi}.
Under the DDH assumption, given encrypted input $\mathcal{E}_{FE}.\texttt{Enc}(\pmb{w}\pmb{x})$ and $\mathcal{E}_{FE}.\texttt{Enc}(\pmb{x})$, there is adversary has non-negligible advantage to break the $\mathcal{E}_{FE}.\texttt{Enc}(\pmb{w}\pmb{x})$ and $\mathcal{E}_{FE}.\texttt{Enc}(\pmb{x})$ to directly obtain $\pmb{w}\pmb{x}$ and $\pmb{x}$, respectively.

\begin{lemma}[Solution of a Non-Homogeneous System]
\label{lemma:equation}
A non-homogeneous system is a linear system of equations $\pmb{A}\pmb{x}^{\intercal}=\pmb{b}$ s.t. $\pmb{b}\ne 0$, where $\pmb{A}\in\mathbb{R}^{m\times n}, \pmb{b}, \pmb{x}\in\mathbb{R}^{n}$.
$\pmb{A}\pmb{x}^{\intercal}=\pmb{b}$ is consistent if and only if the rank of the coefficient matrix $\texttt{rank}(\pmb{A})$ is equal to the rank of the augmented matrix $\texttt{rank}(\pmb{A};\pmb{b})$, while $\pmb{A}\pmb{x}^{\intercal}=\pmb{b}$ has only one solution if and only if $\texttt{rank}(\pmb{A})=\texttt{rank}(\pmb{A};\pmb{b})=n$.  
\end{lemma}

\begin{lemma}[Security and Randomness of OTP-based Seed Generation]
\label{lemma:selection}
Given a predefined party group $\mathcal{P}$ with OTP setting, we have the following claims:
\textsc{Security-}except for the released seeds, $\forall p^{'} \notin \mathcal{P}$, $p^{'}$ cannot infer the next seed based on released seeds;
\textsc{Randomness-}$\forall p_{i} \in \mathcal{P}$, $p_i$ can obtain a synchronized and sequence-related one-time seed without peer-to-peer communication with other parties.
\end{lemma}

The \textsc{Lemma}~\ref{lemma:equation} is derived from the conclusion of the \textit{Rouché-Capelli} theorem \cite{shafarevich2012linear}. Hence, we do not present the specific proof here to avoid redundancy.
The proof of \textsc{Lemma}~\ref{lemma:selection} is presented in Appendix~\ref{sec:app:lemma}.
Based on the above-introduced \textsc{Lemma}~\ref{lemma:security}, \ref{lemma:equation} and \ref{lemma:selection}, we obtain the theorem to claim the privacy guarantee of \textit{FedV} and corresponding proof as follows.

\begin{theorem}[Privacy Guarantee of FedV]
\label{theorem:privacy}
Under the threat models defined in Section~\ref{sec:vfl:threat}, \textit{FedV} can protect the privacy of the parties' input under the inference attack in Definition~\ref{def:infer_attack}.
\end{theorem}

\begin{proof}[Proof of Theorem~\ref{theorem:privacy}]
We prove the theorem by hybrid games to simulate the inference activities of a PPT adversary $\mathcal{A}$. 
\begin{itemize}
    \item[$G_0$:] $\mathcal{A}$ obtains an encrypted input $\texttt{Enc}(\pmb{x})$ to infer $\pmb{x}$;
    \item[$G_1$:] $\mathcal{A}$ observes the randomness of one round of batch selection to infer the next round of batch selection;
    \item[$G_2$:] $\mathcal{A}$ collects a triad of encrypted input, aggregation weight and inner-product, $(\texttt{Enc}(\pmb{x}), \pmb{w}, \langle\pmb{w},\pmb{x}\rangle)$, to infer $\pmb{x}$, $\pmb{w},\pmb{x}\in\mathbb{R}^{n}$;
    \item[$G_3$:] $\mathcal{A}$ collects a set $\mathcal{S}_{(\texttt{Enc}(\pmb{x}_{\text{target}}), \pmb{w} \langle\pmb{w},\pmb{x}_{\text{target}}\rangle)}$ to infer $\pmb{x}_{\text{target}}$.
\end{itemize}
Here, we analyze each inference game and the hybrid cases.
According to \textsc{Lemma}~\ref{lemma:security}, $\mathcal{A}$ does not have non-negligible advantage to infer $\pmb{x}$ by breaking $\texttt{Enc}(\pmb{x})$. 
As we have proved in \textsc{Lemma}~\ref{lemma:selection}, in game $G_1$, $\mathcal{A}$ also does not have non-negligible advantage to infer the next round of batch selection. 
Here, the combination of game $G_0$ or $G_1$ with other games does not increase the advantage of $\mathcal{A}$.

In game $G_2$, suppose that $\mathcal{A}$ has a negligible advantage to infer $\pmb{x}$. Then, $G_2$ can be reduced to that $\mathcal{A}$ has a negligible advantage to solve a non-homogeneous system, $\pmb{w}^{\intercal}\pmb{x}=b$.
Here we consider three cases:

\noindent\textbf{Case $C1$}: Except for directly solving the $\pmb{w}^{\intercal}\pmb{x}=b$ system, $\mathcal{A}$ has no extra ability. According to \textsc{Lemma}~\ref{lemma:equation}, if $\pmb{w}^{\intercal}\pmb{x}=b$ has one confirmed solution, it requires that $n=1$. In \textit{FedV}, the number of features and the batch size setting are grater than one. Thus, $\mathcal{A}$ cannot solve the non-homogeneous system.

\noindent\textbf{Case $C2$}: Based on $C1$, $\mathcal{A}$ could be an aggregator, where $\mathcal{A}$ can manipulate a weight vector $\pmb{w}_{\text{exploited}}$ s.t. $w_i = 1, \forall j\in[n], j\ne i, w_j=0$ to infer $x_i\in\pmb{x}$. However, \textit{FedV} does not allow the functional key generation using $\pmb{w}_{\text{exploited}}$ due to IPM setting. Without functional decryption key, $\mathcal{A}$ cannot acquire the inner-product, i.e., $b$ in the non-homogeneous system. In this case, $\pmb{w}_{\text{exploited}}^{\intercal}\pmb{x}=b$ has multiple solutions, and hence $x_i$ cannot be confirmed. 

\noindent\textbf{Case $C3$}: Based on $C1$, $\mathcal{A}$ could be a group of colluding parties, where $\mathcal{A}$ also have learned part of information of $\pmb{x}$.
Then, the inference task is reduced to solve $\pmb{w}^{'\intercal}\pmb{x}^{'}=b^{'}$ system. According to \textsc{Lemma}~\ref{lemma:equation}, it requires that $|\pmb{x}^{'}|=1$ to have one solution if and only if colluding parties learn $\pmb{w}^{'}$.
In the threat model of \textit{FedV}, aggregator is assumed not colluding with parties in the aggregation process, and hence such a condition is not satisfied.
Thus, $\mathcal{A}$ cannot solve $\pmb{w}^{'\intercal}\pmb{x}^{'}=b^{'}$ system.

In short, $\mathcal{A}$ cannot solve the non-homogeneous system and hence $\mathcal{A}$ does not have a non-negligible advantage to infer $\pmb{x}$ in game $G_2$.

Game $G_3$ is a variant of game $G_2$, where $\mathcal{A}$ collects a set of triads as shown in game $G_2$ for a target data sample $\pmb{x}_{\text{target}}$.
With enough triads, $\mathcal{A}$ can reduce the inference task to the task of constructing a non-homogeneous system of $\pmb{W}\pmb{x}_{\text{target}}^{\intercal}=\pmb{b}$, s.t., $\texttt{rank}(\pmb{W})=\texttt{rank}(\pmb{W};\pmb{b})=n$ as illustrated in  \textsc{Lemma}~\ref{lemma:equation}.
Here we also consider two cases: (i) $\mathcal{A}$ could be the aggregator, however, \textit{FedV} employs the OTP-based seed generation mechanism to chose the samples for each training batch. 
According to game $G_1$, $\mathcal{A}$ does not have a non-negligible advantage to observer and infer random batch selection. 
(ii) $\mathcal{A}$ could be the colluding parties, then it is reduced to $G_2$ case $C3$.
As a result, $\mathcal{A}$ still cannot construct a non-homogeneous system to solve $\pmb{x}_{\text{target}}$.

Based on the above simulation games, $\mathcal{A}$ does not have the non-negligible advantage to infer the private information defined in Definition~\ref{def:infer_attack}
Thus, the privacy guarantee of \textit{FedV} is proved.
\end{proof}

\noindent\textbf{Remark}.
According to our threat model and \textit{FedV} design, labels are kept fully private for linear models by encrypting them during the feature dimension secure aggregation
(Procedure \ref{alg:sg} line \ref{alg:sg:ln:fsa:active}).
For non-linear models, a slightly different process is involved. In this case,
the active party shares the label with the aggregator to avoid costly peer-to-peer communication.
Sharing labels, in this case, does not compromise the privacy of the features of other parties for two reasons.
First, all the features are still encrypted using the feature dimension scheme. Secondly, because the aggregator does not know what samples are involved in each batch (OTP-based seed generation induced randomness and security), it cannot perform either of the previous inference attacks. 

In conclusion, \textit{FedV} protects the privacy of the features provided by all parties.

\section{Evaluation}

\label{sec:eval}

To evaluate the performance of our proposed framework, we compare \textit{FedV} with the following baselines:

\noindent (\romannumeral1) \textit{Hardy}: we use the VFL proposed in \cite{hardy2017private} as the baseline because it is the closest state-of-the-art approach. In \cite{hardy2017private}, the trained ML model is a logistic regression (LR) and its secure protocols are built using additive homomorphic encryption (HE).
Like most of the additive HE based privacy-preserving ML solutions,
the SGD and loss computation in \cite{hardy2017private} relies on the Taylor series expansion to approximately compute the logistic function.

\noindent (\romannumeral2) \textit{Centralized baselines}:
we refer to the training of different ML models in a centralized manner as the \textit{centralized baselines}.
We train multiple models including an LR model with and without Taylor approximation, a basic linear regression model with mean squared loss and a linear Support Vector Machine (SVM).

\noindent\textbf{Theoretical Communication Comparison.}
Before presenting the experimental evaluation, we first theoretically 
compare the number of \textit{communications} between
the proposed \textit{FedV} with respect to \textit{Hardy}.
Suppose that there are $n$ parties and one aggregator in the VFL framework.
As shown in \tablename\;\ref{tab:efficiency}, in total, \textit{FedV} reduces the number of communications during the training process from $4n-2$ for \cite{hardy2017private} to $n$,
while reducing the number of communications during the loss computation (see Appendix~\ref{sec:app:loss} for details)  from $(n^2-3n)/2$ to $n$.
In \textit{FedV}, the number of communications and loss computation phase is linear to the the number of parties.
 
\begin{table}
  \caption{Number of required crypto-related communication for each iteration in the VFL.}
  \label{tab:efficiency}
  \vspace{-3mm}
  \footnotesize
  \begin{threeparttable}
  \begin{tabular}{ccc}
    \toprule
    Communication & Hardy et al.\cite{hardy2017private} & \textit{FedV} \\
    \midrule
    Secure Stochastic Gradient Descent &  & \\
    aggregator $\leftrightarrow$ parties & $2n$ & $n$\\
    parties $\leftrightarrow$ parties & $2(n-1)$ & $0$\\
    TOTAL & $2(2n-1)$ & $n$ \\
    \hline
    Secure Loss Computation &  & \\
    aggregator $\leftrightarrow$ parties & $2n$ & $n$\\
    parties $\leftrightarrow$ parties & $n(n-1)/2$ & 0\\
    TOTAL & $(n^2+3n)/2$ & $n$ \\
    \bottomrule
    \end{tabular}
   \end{threeparttable}
    \vspace{-5mm}
\end{table}

\subsection{Experimental Setup}
To evaluate the performance of \textit{FedV}, we train several popular ML models including linear regression, logistic regression, Taylor approximation based logistic regression, and linear SVM to classify several publicly available datasets from \textit{UCI Machine Learning Repository} \cite{dua2019uci}, including
website \textit{phishing},
\textit{ionosphere},
landsat satellite,
optical recognition of handwritten digits (\textit{optdigits}),
and \textit{MNIST} \cite{lecun2010mnist}.
Each dataset is partitioned vertically and equally according to the numbers of parties in all experiments. 
The number of attributes of these datasets is between 10 and 784, while the total number of sample instances is between 351 and 70000, and the details can be found in \tablename~\ref{tab:dataset} of Appendix~\ref{sec:app:data}.
Note that we use the same underlying logic used by the popular \textit{Scikit-learn ML} library to handle multi-class classification models, we convert the multi-label datasets into binary label datasets, which is also the strategy used in the comparable literature \cite{hardy2017private}.

\begin{figure*}[t] 
	\centering 
	\includegraphics[width=\linewidth,trim=0 0 120 15,clip]{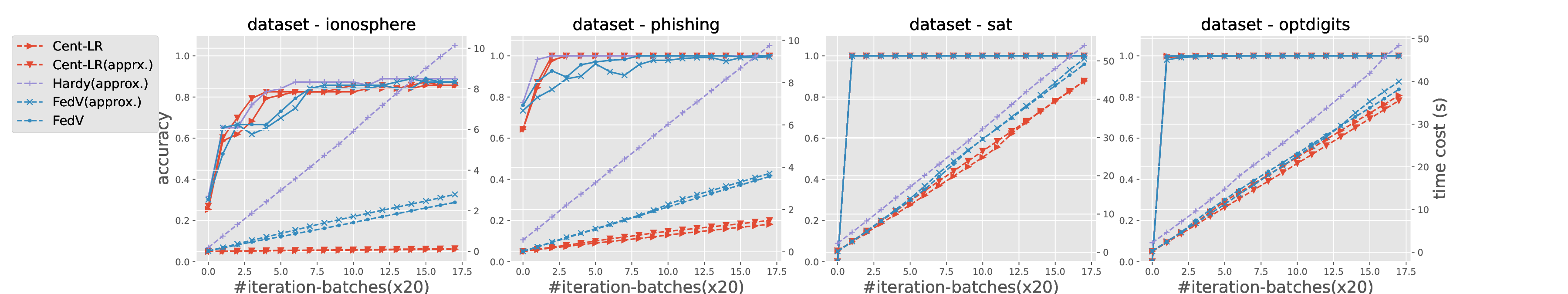}
	\vspace{-2mm}
	\caption{Model accuracy and training time comparisons for logistic regression with two parties. The accuracy and training time is presented in the first and second rows. Each column presents the results for different datasets.}
	\label{fig:cmp:baselines} 
	\vspace{-0.5cm}
\end{figure*}

\noindent\textit{\textbf{Implementation}}.
We implemented \textit{Hardy}, our proposed \textit{FedV} and several centralized baseline ML models in Python.
To achieve the integer group computation that is required by both the additive homomorphic encryption and the functional encryption, we employ the \textit{gmpy2} library \footnote{https://pypi.org/project/gmpy2/}.
We implement the Paillier cryptosystem for the construction of an additive HE scheme; this is the same as the one used in \cite{hardy2017private}. The constructions of MIFE and SIFE are from \cite{abdalla2015simple} and \cite{abdalla2018multi}, respectively.
As these constructions do not provide the solution to address the discrete logarithm problem in the decryption phases, which is a performance intensive computation, we use the same hybrid approach that was used in \cite{xu2019hybridalpha}.
Specifically, to compute $f$ in $h=g^f$, we setup a hash table $T_{h,g,b}$ to store $(h, f)$ with a specified $g$ and a bound $b$, where $ -b \le f \le b$, when the system initializes. 
When computing discrete logarithms, the algorithm first looks up $T_{h,g,b}$ to find $f$, the complexity for which is $\mathcal{O}(1)$.
If there is no result in $T_{h,g,b}$, the algorithm employs the traditional \textit{baby-step giant-step} algorithm \cite{shanks1971class} to compute $f$, the complexity for which is $\mathcal{O}(n^{\frac{1}{2}})$.

\noindent\textit{\textbf{Experimental Environment}}.
All the experiments are performed on a 2.3 GHz 8-Core Intel Core i9 platform with 32 GB of RAM. Both \textit{Hardy} and our \textit{FedV} frameworks are distributed among multiple processes, where each process represents a party.
The parties and the aggregator communicate using local sockets;
hence the network latency is not measured in our experiment.

\begin{figure}[h!] 
	\centering 
 	\includegraphics[width=0.45\textwidth,trim=30 0 40 20,clip]{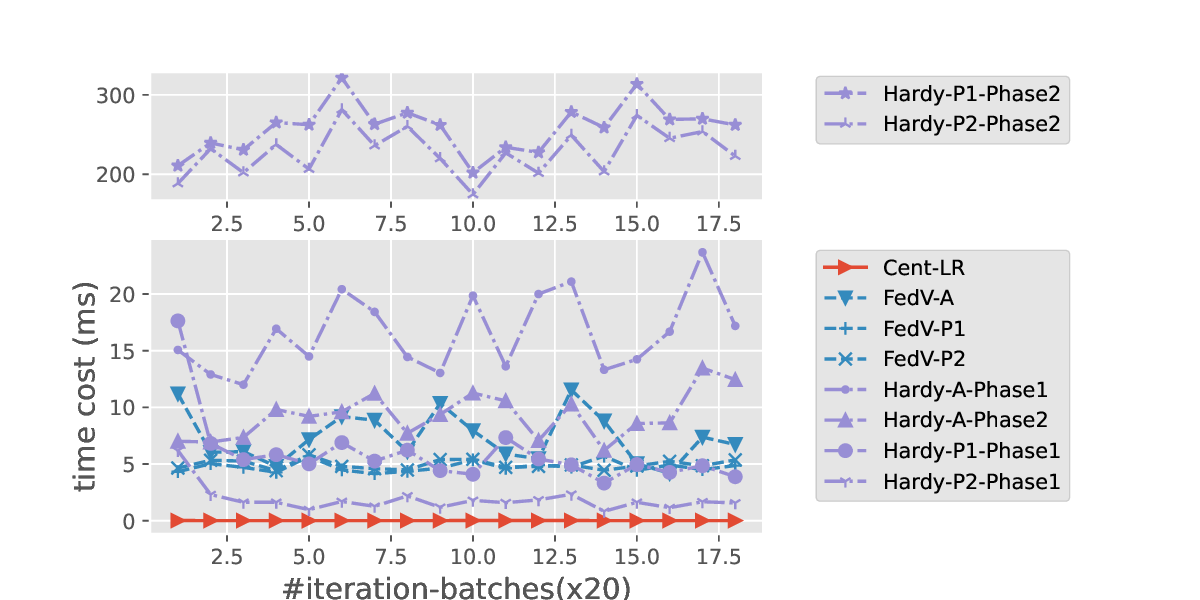}
 	\vspace{-2mm}
	\caption{Decomposition of training time. In the legend, ``A'' represents the aggregator, while ``P1'' and ``P2'' denote the active party and the passive party, respectively.}
	\label{fig:cmp:training_time:spec} 
	\vspace{-0.3cm}
\end{figure}

\begin{figure}[h!] 
	\centering 
	\includegraphics[width=0.3\textwidth,trim=10 0 20 10,clip]{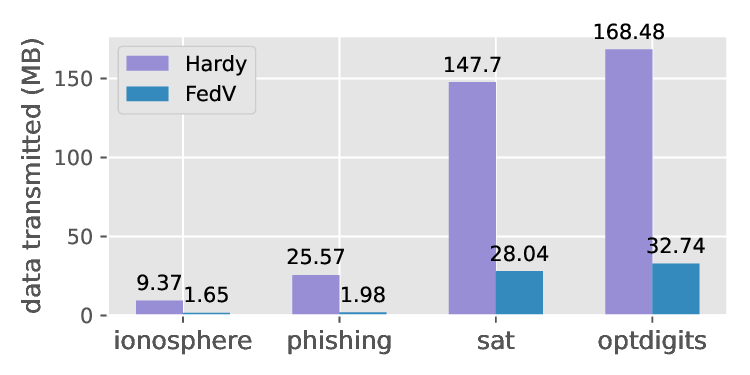}
	\vspace{-2mm}
	\caption{Total data transmitted while training a LR model over 20 training epochs with two parties}
	\label{fig:cmp:transmitted:spec} 
	\vspace{-0.5cm}
\end{figure}

\subsection{Experimental Results}

As \textit{Hardy} only supports two parties to train a logistic regression model, we first present the comparison results for that setting. Then, we explore the performance of \textit{FedV} using different ML models. Lastly, we study the impact of varying number of parties in \textit{FedV}.

\noindent\textbf{Performance of FedV for Logistic Regression.}
We trained two models with \textit{FedV}:
1) a logistic regression model trained according to Procedure~\ref{alg:sg:non-liner}, referred as \textit{FedV};
and 2) a logistic regression model with Taylor series approximation, which reduces the logistic regression model to a linear model, trained according to Procedure~\ref{alg:sg} and
referred as \textit{FedV with approximation}.
We also trained a centralized version (non-FL setting) of a logistic regression with and without Taylor series approximation, referred as \textit{centralized LR} and \textit{centralized LR (approx.)}, respectively. We also present the results for \textit{Hardy}.


\figurename~\ref{fig:cmp:baselines} shows the test accuracy and training time of each approach to train the logistic regression on different datasets.
Results show that 
both of our \textit{FedV} and \textit{FedV with approximation} can achieve a test accuracy comparable to those of the \textit{Hardy} and the \textit{centralized baselines} for all four datasets.
With regards to the training time, \textit{FedV} and \textit{FedV with approximation}  efficiently reduce the training time by 10\% to 70\% for the chosen datasets with $360$ total training epochs.
For instance, as depicted in \figurename~\ref{fig:cmp:baselines}, \textit{FedV} can reduce around 70\% training time for the \textit{ionosphere} dataset while reducing around 10\% training time for the \textit{sat} dataset. 
The variation in training time reduction among different datasets is caused by different data sample sizes and model convergence speed.

\begin{figure*}[h!] 
	\centering 
	\includegraphics[width=\linewidth,trim=0 0 120 13,clip]{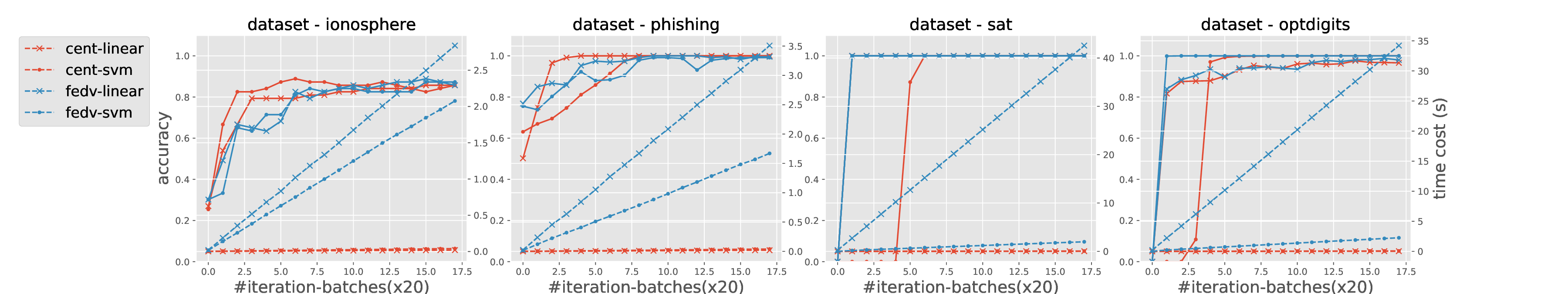}
	\vspace{-2mm}
	\caption{Accuracy and training time during training linear regression and linear SVM for two-party setting. Columns show the results for different datasets.}
	\label{fig:cmp:model} 
	\vspace{-5mm}
\end{figure*}

We decompose the training time required to train the LR model to understand the exact reason for such reduction. These results are shown for the \textit{ionosphere} dataset.
In \figurename~\ref{fig:cmp:training_time:spec}, we can observe that \textit{Hardy} requires
communication between parties and the aggregator (phase 1) and peer-to-peer communication (phase 2).
In contrast, \textit{FedV} does not require peer-to-peer communication, resulting in savings in training times.
Additionally, it can be seen that the computational time for phase 1 of the aggregator and phase 2 of each party are significantly higher for \textit{Hardy} than for \textit{FedV}.
We also compare and decompose the total size of data transmitted for the LR model over various datasets.
As shown in \figurename~\ref{fig:cmp:transmitted:spec}, compared to \textit{Hardy}, \textit{FedV} can reduce the total amount of data transmitted by 80\% to 90\%; this is possible because \textit{FedV} only relies on non-interactive secure aggregation protocols and does not need the frequent rounds of communications used by the contrasted VFL baseline.

\noindent\textbf{Performance of FedV with Different ML Models.}
We explore the performance of \textit{FedV} using various popular ML models including linear regression and linear SVM. 

The first row of \figurename~\ref{fig:cmp:model} shows the test accuracy while the second row shows the training time for a total of $360$ training epochs.
In general, our proposed \textit{FedV} achieves comparable test accuracy for all types of ML models for the chosen datasets.
Note that our \textit{FedV} is based on cryptosystems that compute over integers instead of floating-point numbers, so as expected, \textit{FedV} will lose a portion of fractional parts of a floating-point numbers.
This is responsible for the differences in accuracy with respect to the central baselines.
As expected, compared with our centralized baselines, \textit{FedV} requires more training time. This is due to the distributed nature of the vertical training process.

\begin{figure}[h!] 
	\centering 
	
	\begin{subfigure}[]{0.45\textwidth}
        \centering
        \includegraphics[width=.95\textwidth,trim=5 5 5 0,clip]{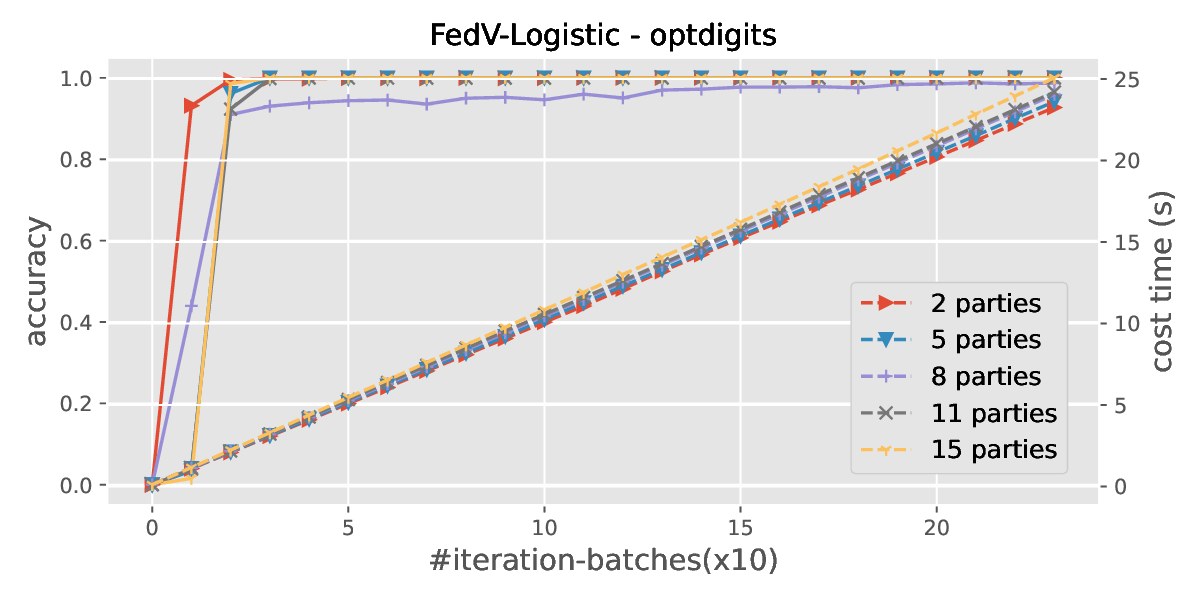}
        \vspace{-2mm}
        \caption{Impact of increasing parties \# on the performance of the \textit{FedV}. \textit{Hardy} \cite{hardy2017private} only works for two parties, hence its not included in the figure.}
	\label{fig:cmp:multiple} 
    \end{subfigure}
    \hfill
    \begin{subfigure}[]{0.45\textwidth}
        \centering
        \includegraphics[width=.95\textwidth,trim=5 5 5 0,clip]{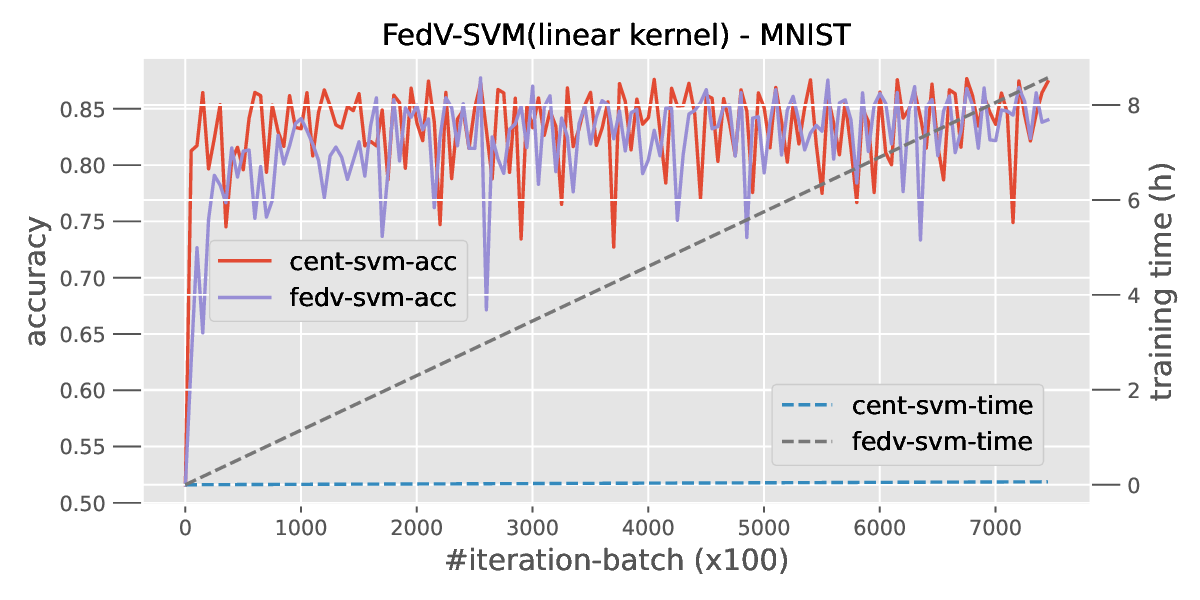}
        \vspace{-2mm}
        \caption{Accuracy and training time during training SVM with linear kernel on MNIST dataset. Hardy was not proposed for this type of model; hence it is not shown.}
	\label{fig:cmp:mnist} 
    \end{subfigure}
	\caption{Performance for Image dataset}
	\vspace{-5mm}
\end{figure}

\noindent\textbf{Impact of Increasing the Number of Parties.}
We explore the impact of increasing number of parties in \textit{FedV}.
Recall that \textit{Hardy} does not support more than two parties, and hence we cannot report its performance in this experiment.
\figurename~\ref{fig:cmp:multiple} shows the accuracy and training time of \textit{FedV} for collaborations varying from two to 15 parties.
The results are shown for the \textit{OptDigits} dataset and the trained model is a Logistic Regression. 

As shown in \figurename~\ref{fig:cmp:multiple}, the number of parties does not impact the model accuracy and finally all test cases reach the 100\% accuracy.
Importantly, the training time shows a linear relation to the number of parties.
As reported in \figurename~\ref{fig:cmp:baselines}, the training time of \textit{FedV} in logistic regression model is very close to that of the normal non-FL logistic regression.
For instance, for 100 iterations, the training time for \textit{FedV} with 14 parties is around 10 seconds, while the training time for normal non-FL logistic regression is about 9.5 seconds.
We expect this time will increase in a fully distributed setting depending on the latency of the network.

\noindent\textbf{Performance on Image Dataset.} 
\figurename~\ref{fig:cmp:mnist} reports the training time and model accuracy for training a linear SVM model on MNIST dataset using a batch size of 8 for 100 epochs. Note that \textit{Hardy} is not reported here because that approach was proposed for approximated logistic regression model, but not for linear SVM.
Compared to the centralized linear SVM model, \textit{FedV} can achieve comparable model accuracy.
While \textit{FedV} provides a strong security guarantee, the training time is still acceptable.

Overall, our experiments show reductions of 10\%-70\% of training time and 80\%-90\% transmitted data size compared to \textit{Hardy}.
We also showed that \textit{FedV} is able to
train machine learning models that the baseline cannot train (see Figure \ref{fig:cmp:mnist}). \textit{FedV}
final model accuracy was comparable to central baselines showing the advantages of not requiring Taylor approximation techniques used by \textit{Hardy}.

\section{Related Work}
\label{sec:related}

FL was proposed in \cite{mcmahan2016communication, konevcny2016federated} to allow a group of collaborating parties to jointly learn a global model without sharing their data \cite{li2019federated}.
Most of the existing work in the literature focus on horizontal FL while these papers address issues related to privacy and security \cite{bonawitz2017practical,geyer2017differentially, xu2019hybridalpha, truex2019hybrid,bagdasaryan2018backdoor,nasr2019comprehensive, fredrikson2015model,shokri2017membership, nasr2018machine},
system architecture \cite{ludwig2020ibm,konevcny2016federated, mcmahan2016communication, bonawitz2019towards}, and new learning algorithms e.g., \cite{zhao2018federated, smith2017federated, corinzia2019variational}.


A few existing approaches have been proposed for distributed data mining
\cite{vaidya2008survey,yu2006privacy,slavkovic2007secure,vaidya2008privacy}. A survey of vertical data mining methods is presented in \cite{vaidya2008survey}, where these methods are proposed to train specific ML models such as support vector machine \cite{yu2006privacy}, logistic regression \cite{slavkovic2007secure} and decision tree \cite{vaidya2008privacy}. These solutions are not designed to prevent inference/privacy attacks.
For instance, in \cite{yu2006privacy}, the parties form a ring where a first party adds a random number to its input and sends it to the following one; each party adds its value and sends to the next one; and the last party sends the accumulated value to the first one. Finally, the first party removes the random number and 
broadcasts the aggregated results. Here, it is possible to infer private information given that each party knows intermediate and final results.
The privacy-preserving decision tree model in \cite{vaidya2008privacy} has to reveal class distribution over the given attributes, and thus may have privacy leakage.
Split learning \cite{vepakomma2018split,singh2019detailed}, a new type of distributed deep learning, was recently proposed to train neural networks without sharing raw data. Although it is mentioned that secure aggregation may be incorporated during the method, no discussion on the possible cryptographic techniques were provided.
For instance, it is not clear if the an applicable cryptosystem would require Taylor approximation.
None of these approaches provide strong privacy protection against inference threats.


Some proposed approaches have incorporated privacy into vertical FL \cite{gascon2016secure,hardy2017private,cheng2019secureboost,yang2019quasi,gu2020federated,zhang2021secure,chen2020vafl,wang2020hybrid}.
These approaches are either limited to a specific model type: 
a procedure to train secure linear regression was presented in \cite{gascon2016secure},
a private logistic regression process was presented in \cite{hardy2017private}, 
and \cite{cheng2019secureboost} presented an approach to train XGBoost models,
or suffered from the privacy, utility and communication efficiency concerns:
differential privacy based noise perturbation solutions were presented in \cite{chen2020vafl,wang2020hybrid},
random noise perturbation with tree-based communication were presented in \cite{gu2020federated, zhang2021secure}.
There are several differences between these approaches and \textit{FedV}.
First, these solutions either rely on the hybrid general (garbled circuit based) secure multi-party computation approach or are built on partially additive homomorphic encryption (i.e., Paillier cryptosystem \cite{damgaard2001generalisation}).
In these approaches, the secure aggregation process is inefficient in terms of communication and computation costs compared to our proposed approach (see Table \ref{tab:efficiency}).
Secondly, they also require approximate computation for non-linear ML models (Taylor approximation); this results in lower model performance compared to the proposed approach in this paper.
Finally, they increase the communication complexity or reduce the utility since the noise perturbation is introduced in the model update.

The closest approach to \textit{FedV} is \cite{hardy2017private,yang2019quasi}, which makes use of Pailler cryptosystem and only supports linear-models;
a detailed comparison is presented in our experimental section.
The key differences between the two approaches are as follows:
(i) \textit{FedV} does not require any peer-to-peer communication; as a result the training time is drastically reduced as compared to the approach in \cite{hardy2017private,yang2019quasi};
(ii), \textit{FedV} does not require the use of Taylor approximation; this results in higher model performance in terms of accuracy; and (iii) \textit{FedV} is applicable for both linear and non-linear models, while the approach in \cite{hardy2017private, yang2019quasi} is limited to logistic regression only. 


Finally, multiple cryptographic approaches have been proposed for secure aggregation, including (\romannumeral1) general secure multi-party computation techniques \cite{huang2011faster, wang2017global, wang2017authenticated, chase2019reusable} that are built on the garbled circuits and oblivious transfer techniques; (\romannumeral2) secure computation using more recent cryptographic approaches such as homomorphic encryption and its variants \cite{lopez2012fly, damgaard2012multiparty, baum2016better, keller2018overdrive, araki2018choose}.
However, these two kinds of secure computation solutions have limitations with regards to either the large volumes of ciphertexts that need to be transferred or the inefficiency of computations involved (i.e., unacceptable computation time).
Furthermore, to lower communication overhead and computation cost, customized secure aggregation approaches such as the one proposed in \cite{bonawitz2017practical} are mainly based on secret sharing techniques and they use authenticated encryption to securely compute sums of vectors in horizontal FL. In 
\cite{xu2019hybridalpha}, Xu et al. proposed the use of functional encryption \cite{lewko2010fully, boneh2011functional} to enable \textit{horizontal} FL. However,
this approach cannot be used to handle the secure aggregation requirements in \textit{vertical} FL.


%
\section{Conclusions}
\label{sec:conclusion}
Most of the existing privacy-preserving FL frameworks only focus on horizontally partitioned datasets. The few existing vertical federated learning solutions work only on a specific ML model and suffer from inefficiency with regards to secure computations and communications.
To address the above-mentioned challenges,
we have proposed \textit{FedV}, an efficient and privacy-preserving VFL framework based on a two-phase non-interactive secure aggregation approach that makes use of functional encryption.

We have shown that \textit{FedV} can be used to train a variety of ML models, without a need for any approximation, including logistic regression,
SVMs, among others.
\textit{FedV} is the first VFL framework that supports parties to dynamically drop and re-join for all these models during a training phase; thus, it is applicable in challenging situations where a party may be unable to sustain connectivity throughout the training process.
More importantly, \textit{FedV} removes the need of peer-to-peer communications among parties, thus, reducing substantially the training time and making it applicable to applications where parties cannot connect with each other.
Our experiments show reductions of 10\%-70\% of training time and 80\%-90\% transmitted data size compared to those in the state-of-the art approaches.


\bibliographystyle{plain}
\bibliography{references}

\appendix

\section{Formal Analysis of Claim \ref{theo:gp_vfl_theorem}}
\label{appendix:proof}

Here, we present our detailed proof of Claim \ref{theo:gp_vfl_theorem}.
Note that we skip the discussion on how to compute $\nabla R$ in the rest of the analysis such as in equation \eqref{equ:linear_gradient} below, since the aggregator can compute it independently.

\subsection{Linear Models in \textit{FedV}}
Here, we formally analyze the details of how our proposed \textit{Fed-SecGrad} approach (also called \textit{2Phase-SA}) is applied in a vertical federated learning framework with underlying linear ML model. Suppose the a generic \textit{linear model} is defined as:
\begin{equation}
\label{equ:fit}
f(\pmb{x};\pmb{w}) = w_{0}x_0 + w_{1}x_{1} + ... + w_{d}x_{d}, 
\end{equation}
where $x^{(i)}_0 =1$ represents the bias term. 
For simplicity, we use the vector-format expression in the rest of the proof, described as: $f(\pmb{x};\pmb{w}) = \pmb{w}^{\intercal}\pmb{x}$, where $\pmb{x} \in \mathbb{R}^{d+1}, \pmb{w} \in \mathbb{R}^{d+1}, x_0 =1$.
Note that we omit the bias item $w_0x_0=w_0$ in the rest of analysis as the aggregator can compute it independently.
Suppose that the loss function here is least-squared function, defined as
\begin{equation}
\label{equ:loss}
\mathcal{L}(f(\pmb{x};\pmb{w}), y) = \tfrac{1}{2}(f(\pmb{x};\pmb{w}) - y)^2
\end{equation}
and we use L2-norm as the regularization term, defined as $R(\pmb{w}) = \frac{1}{2}\sum^{n}_{i=1}w^2_i$.
According to equations \eqref{equ:train_loss}, \eqref{equ:fit} and \eqref{equ:loss}, the gradient of $E(\pmb{w})$ computed over a mini-batch $\mathcal{B}$ of $s$ data samples is as follows:

\begin{align}
\nabla E_{\mathcal{B}}(\pmb{w}) 
&= \nabla \mathcal{L}_{\mathcal{B}}(\pmb{w}) + \nabla R_{\mathcal{B}}(\pmb{w})\nonumber\\
&=\tfrac{1}{s}\tsum^{s}_{i=1}(\pmb{w}^{\intercal}\pmb{x}^{(i)}-y^{(i)})\pmb{x}^{(i)} + \nabla R_{\mathcal{B}}(\pmb{w}) 
\label{equ:linear_gradient}
\end{align}

Suppose that $p_1$ is the \textit{active party} with labels $y$, then secure gradient computation can be expressed as follows:
\begin{align}
\nabla E 
&= \tfrac{1}{s}\tsum^{s}_{i=1}(
\underbrace{w_{1}x^{(i)}_{1} -y^{(i)}}_{u^{(i)}_{p_1}} + ... + \underbrace{w_{d}x^{(i)}_{d}}_{u^{(i)}_{p_n}})\pmb{x}^{(i)}\\
&= \tfrac{1}{s}\tsum^{s}_{i=1}\underbrace{\tsum^{n}_{j=1}(u^{(i)}_{p_j})}_{\text{feature dimension SA}}\pmb{x}^{(i)}
\end{align}
Next, let $u^{(i)}$ be the intermediate value to represent the \textit{difference-loss} for current $\pmb{w}$ over one sample $\pmb{x}^{(i)}$, which is also the aggregation result of \textit{feature dimension SA}.
Then, the updated gradient $\nabla E_{\mathcal{B}}(\pmb{w})$ is continually computed as follows:
\begin{align}
\nabla E 
&=\tfrac{1}{s}\tsum^{s}_{i=1}u^{(i)}(
\underbrace{x^{(i)}_{1}}_{p_{1}}, ..., 
\underbrace{x^{(i)}_{d}}_{p_{n}}) \\
&=
\underbrace{\tfrac{1}{s}\tsum^{n}_{i=1}u^{(i)}x^{(i)}_{1}}_{{\text {sample dimension SA}}},  ...,
\underbrace{\tfrac{1}{s}\tsum^{n}_{i=1}u^{(i)}x^{(i)}_{d}}_{{\text{sample dimension SA}}}
\end{align}

To deal with the secure computation task of training loss as described in Algorithm \ref{alg:vfl}, we only apply \textit{feature dimension SA} approach.
As the average loss function here is least-squares function, secure computation involved is as follows:
\begin{equation}
\begin{split}
\mathcal{L}_{\mathcal{B}}(\pmb{w}) &= \tfrac{1}{s}\underbrace{\tsum^{s}_{i=1}(\underbrace{\pmb{w}^{\intercal}\pmb{x}^{(i)} - y^{(i)}}_{\text{feature dimension SA}})^2}_{\text{Normal Computation}}
\end{split}
\end{equation}
Obviously, the \textit{feature dimension SA} can satisfy the computation task in the above equation.

\subsection{Generalized Linear Models in \textit{FedV}}

Here we formally analyze the details of applying our \textit{FedV-SecGrad} approach to train generalized linear models in \textit{FedV}.

We use \textit{logistic regression} as an example, which has the following fitting (prediction) function:
\begin{equation}
    \label{equ:lr_f}
    f(\pmb{x};\pmb{w}) = \tfrac{1}{1+e^{-\pmb{w}^{\intercal}\pmb{x}}}
\end{equation}
For binary label $y \in \{0,1\}$, the loss function can be defined as:
\begin{equation}
    \label{equ:lr_loss}
    \mathcal{L}(f(\pmb{x};\pmb{w}), y) = \left\{
    \begin{array}{l}
        -\log(f(\pmb{x};\pmb{w}))\; \text{if } y=1\\
        -\log(1-f(\pmb{x};\pmb{w}))\;\text{if } y=0
    \end{array}\right.
\end{equation}
Then, the total loss over a mini-batch $\mathcal{B}$ of size $s$ is computed as follows:
\begin{align}
E(\pmb{w}) 
& = -\tfrac{1}{s}\tsum^{s}_{i=1}[y^{(i)}\pmb{w}^{\intercal}\pmb{x}^{(i)} - \log(1+e^{\pmb{w}^{\intercal}\pmb{x}^{(i)}})]
\end{align}
The gradient is computed as follows:
\begin{align}
\nabla E(\pmb{w})
&= \tfrac{1}{s}\tsum^{s}_{i=1}[\tfrac{\partial}{\partial\pmb{w}}\log(1+e^{\pmb{w}\pmb{x}^{(i)}}) - \tfrac{\partial y^{(i)}\pmb{w}\pmb{x}^{(i)}}{\partial\pmb{w}}]\\
&= \tfrac{1}{s}\tsum^{s}_{i=1}(\frac{1}{1+e^{-\pmb{w}\pmb{x}^{(i)}}} - y^{(i)})\pmb{x}^{(i)}
\end{align}
Note that we also do not include the regularization term $\lambda R(\pmb{w})$ here for the same aforementioned reason.
Here, we show two potential solutions in detail:

\noindent\textit{\textbf{(\romannumeral1)}  \textit{FedV} for nonlinear model (Procedure~\ref{alg:sg:non-liner})}.
Firstly, although the prediction function in (\ref{equ:lr_f}) is a non-linear function, it can be decomposed as $f(\pmb{x};\pmb{w})=g(h(\pmb{x};\pmb{w}))$, where:
\begin{equation}
g(h(\pmb{x};\pmb{w}))=\tfrac{1}{1+e^{-h(\pmb{x};\pmb{w})}} \text{, where } h(\pmb{x};\pmb{w}) = \pmb{w}^{\intercal}\pmb{x}
\end{equation}
We can see that the sigmoid function $g(z)=\frac{1}{1+e^{-z}}$ is not a linear function, while $h(\pmb{x};\pmb{w})$ is linear.
We then apply our feature dimension and sample dimension secure aggregations on linear function $h(\pmb{x};\pmb{w})$ instead.
To be more specific, the formal expression of the secure gradient computation is as follows:
\begin{align}
\nabla E(\pmb{w})
&= \tfrac{1}{s}\tsum^{s}_{i=1}
\underbrace{(\tfrac{1}{1+e^{-\sum^{n}_{j}(u^{(i)}_{p_{j}})\}\rightarrow\text{feature dim. SA}}} - y^{(i)})}_{\text{Normal Computation}\rightarrow u^{(i)} }\pmb{x}^{(i)}\nonumber\\
&=\tfrac{1}{s}(
\underbrace{\tsum^{s}_{i=1}u^{(i)}x^{(i)}_{1}}_{{\text{sample dimension SA}}},  ..., 
\underbrace{\tsum^{s}_{i=1}u^{(i)}x^{(i)}_{j}}_{{\text{sample dimension SA}}})
\end{align}
Note that the output of \textit{feature dimension SA} is in plaintext, and hence, the aggregator is able to evaluate the sigmoid function $g(\cdot)$ together with the labels.
The secure loss can be computed as follows:
\begin{align}
E(\pmb{w}) &= 
-\tfrac{1}{s}\tsum^{s}_{i=1}[
\underbrace{y^{(i)}\underbrace{\pmb{w}^{\intercal}\pmb{x}^{(i)}}_{\text{feature dim. SA}}}_{\text{normal computation}} - \nonumber\\
& \;\;\;\;\;
\underbrace{\log(1+e^{-\pmb{w}^{\intercal}\pmb{x}^{(i)}\}\rightarrow\text{feature dim. SA}})}_{\text{normal computation}}
]
\end{align}
Similar to secure gradient descent computation, however, we only have the \textit{feature dimension SA} with subsequent normal computation.

\noindent\textit{\textbf{(\romannumeral2)} Taylor approximation}.
\textit{FedV} also supports the Taylor approximation approach as proposed in \cite{hardy2017private}.
In this approach, the Taylor series expansion of function $\log(1+e^{-z})$=$\log2-\frac{1}{2}z+\frac{1}{8}z^2+O(z^4)$ is applied to equation (\ref{equ:lr_loss}) to approximately compute the gradient as follows:
\begin{equation}
\label{equ:approx_gradient}
    \nabla E_{\mathcal{B}}(\pmb{w}) \approx \frac{1}{s}\sum^{s}_{i=1}(\frac{1}{4}\pmb{w}^{\intercal}\pmb{x}^{(i)}-y^{(i)}+\frac{1}{2})\pmb{x}^{(i)}
\end{equation}
As in equation (\ref{equ:linear_gradient}), we are able to apply the \textit{2Phase-SA} approach in the secure computation of equation (\ref{equ:approx_gradient}).

We also use another ML model, \textit{SVMs with Kernels}, as an example. 
Here, we consider two cases: 

\noindent(\romannumeral1) \textit{linear SVM model for data} that is not linearly separable: Suppose that the linear SVM model uses squared hinge loss as the loss function, and hence, its objective is to minimize 
\begin{align}
    E(\pmb{w}) = \tfrac{1}{s}\tsum^{s}_{i=1}\left(\max(0, 1-y^{(i)}\pmb{w}\pmb{x}^{(i)})\right)^2
\end{align}
The gradient can be computed as follows:
\begin{align}
    \nabla E = \tfrac{1}{s}\tsum^{s}_{i=1}[
    -2y^{(i)}(\max(0, 1-y^{(i)}\pmb{w}\pmb{x}^{(i)}))\pmb{x}^{(i)}]
\end{align}
As we know, the aggregator can obtain $\pmb{w}\pmb{x}^{(i)}$ via feature dimension SA.
With the provided labels and $\pmb{w}^{(i)\intercal}\pmb{x}^{(i)}$, \textit{FedV-SecGrad} can update Line~\ref{alg:sg:ln:u} of Procedure~\ref{alg:sg} so that the aggregator computes $u_i= -2y^{(i)}\max(0, 1-y^{(i)}\pmb{w}^{(i)\intercal}\pmb{x}^{(i)})$ instead.

\noindent(\romannumeral2) \textit{the case where SVM uses nonlinear kernels}: 
The prediction function is as follows: 
\begin{equation}
 f(\pmb{x};\pmb{w}) = \tsum^{s}_{i=1} w_iy_ik(\pmb{x}_i, \pmb{x}),   
\end{equation}
where $k(\cdot)$ denotes the corresponding kernel function. 
As nonlinear kernel functions, such as polynomial kernel $(\pmb{x}^{\intercal}_i\pmb{x}_j)^d$, sigmoid kernel $tanh(\beta \pmb{x}^{\intercal}_i\pmb{x}_j + \theta)$ ($\beta$ and $\theta$ are kernel coefficients), are based on inner-product computation which is supported by our \textit{feature dimension SA} and \textit{sample dimension SA} protocols, these kernel matrices can be computed before the training process. 
And the aforementioned objective for SVM with nonlinear kernels will be reduced to SVM with linear kernel case with the pre-computed kernel matrix. Then the gradient computation process for these SVM models will be reduced to a gradient computation of a standard linear SVM,
which can clearly be supported by \textit{FedV-SecGrad}.

\section{Secure Loss Computation in \textit{FedV}}
\label{sec:app:loss}
Unlike the \textit{secure loss computation (SLC)} protocol in the contrasted VFL framework \cite{hardy2017private}, the SLC approach in \textit{FedV} is much simpler. Here, we use the logistic regression model as an example.
As illustrated in Procedure~\ref{alg:secure_lc}, unlike the SLC in \cite{hardy2017private} that is separate and different from the secure gradient computation, the SLC here does not need additional operations for the parties.
The loss result is computed by reusing the result of the \textit{feature dimension SA} in the \textit{FedV-SecGrad}.
\LinesNotNumbered
\begin{algorithm}[!t]
    \small
    \caption{FedV-Secure Loss Computation. \newline
    {\small {\bf Premise:} As the procedure inherits from Procedure~\ref{alg:sg}, we omitted same operations.}
    }
    \label{alg:secure_lc}
    \SetKwProg{Fn}{function}{}{}
    \SetKwProg{Aggregator}{Aggregator}{}{}
    \Aggregator{}{ 
        \setcounter{AlgoLine}{11}
        \nl\ForEach{$k \in \{1,...,s\}$}{
            \nl $u_k \leftarrow \mathcal{E}_\text{MIFE}.\text{Dec}_{sk^{\text{MIFE}}_{\pmb{v}}}(\{\mathcal{C}^{\text{fd}}_{i,k}\}_{i\in\{1,...,n\}})$
        }
        \nl $E_{\mathcal{B}}(\pmb{w}) \leftarrow -\tfrac{1}{s}\tsum^{s}_{i=1}[y^{(i)}u_i - \log(1+e^{u_i})]$\;
    }
\end{algorithm}

\section{Proof of Lemma~\ref{lemma:selection}}
\label{sec:app:lemma}

Here, we present the specific proof for Lemma~\ref{lemma:selection}.
Given a predefined party group $\mathcal{P}$ with OTP setting, we have the following claims:
\textsc{Security-}except for the released seeds, $\forall p^{'} \notin \mathcal{P}$, $p^{'}$ cannot infer the next seed based on released seeds;
\textsc{Randomness-}$\forall p_{i} \in \mathcal{P}$, $p_i$ can obtain a synchronized and sequence-related one-time seed without peer-to-peer communication with other parties.

\begin{proof}[Proof of Lemma~\ref{lemma:selection}]
Since there exist various types of OPT, we adopt the hash chains-based OPT to prove the security and randomness of the OTP-based seed generation.
Given the cryptographic hash function $H$, the initial random seed $r$ and a sequence index $b_{i}$ (i.e., the batch index in the \textit{FedV} training), the OTP-based seed for $b_{i}$ is $r_{i}=H^{(t)}(r)$.
Note that $H^{(t)}(r)=H(H^{(t-1)}(r))= ... =H(...H^{(1)}(r))$.
Next, the OTP-based seed for $b_{i+1}$ is $r_{i+1}=H^{(t-1)}(r)$, and hence we have $r_{i}=H(r_{i+1})$.
Given the $r_{i}$ at training index $b_i$, suppose that an adversary has a non-negligible advantage to infer the next $r_{i+1}$, the adversary needs to find a way of calculating the inverse function $H^{-1}$. Since the cryptographic hash function should be one-way and proved to be computationally intractable according to the adopted schemes, the adversary does not have the non-negligible advantage to infer the next seed. 
With respect to the randomness, the initial seed $r$ is randomly selected, and the follow-up computation over $r$ is a sequence of the hash function, which does not break the randomness of $r$. 
\end{proof}

\section{Dataset Description}
\label{sec:app:data}
As shown in \tablename~\ref{tab:dataset}, we present the dataset we used and the division of training set and test set.

\begin{table}
  \caption{Datasets used for the experimental evaluation.}
  \centering
  \label{tab:dataset}
  \footnotesize
  \begin{threeparttable}
  \begin{tabular}{ccccc}
    \toprule
    Dataset & Attributes \# & Total Samples \# & Training \# & Test \#\\ 
    \midrule
    \textit{Phishing} & 10 & 1353 & 1120 & 233 \\
    \textit{Ionosphere} & 34 & 351 & 288 & 63\\
    \textit{Statlog} & 36 & 6435 & 4432 & 2003\\
    \textit{OptDigits}& 64 & 5620 & 3808 & 1812\\
    \textit{MNIST}& 784 & 70000 & 60000 & 10000\\
    \bottomrule
    \end{tabular}
   \end{threeparttable}
\end{table}

\section{Functional Encryption Schemes}
\label{sec:app:fe}

\subsection{Single-input FEIP Construction}
The single-input functional encryption scheme for the inner-product function $f_{\text{SIIP}}(\pmb{x},\pmb{y})$ is defined as follows:
$$\mathcal{E}_{\text{SIFE}}=(\mathcal{E}_{\text{SIFE}}.\text{Setup}, \mathcal{E}_{\text{SIFE}}.\text{DKGen}, \mathcal{E}_{\text{SIFE}}.\text{Enc}, \mathcal{E}_{\text{SIFE}}.\text{Dec}).
$$
Each of the algorithms is constructed as follows:
\begin{itemize}
    \item $\mathcal{E}_{\text{SIFE}}.\text{Setup}(1^{\lambda}, \eta)$: The algorithm first generates two samples as $(\GG, p, g)$ $\sample$ $\textit{GroupGen}(1^{\lambda})$, and $\pmb{s}$ $=$ $(s_1$ $,...,$ $s_{\eta})$ $\sample$ $\ZZ^{\eta}_{p}$ on the inputs of security parameters $\lambda$ and $\eta$,
    and then sets $\texttt{pp} = (g, h_{i} = g^{s_{i}})_{i\in [1, ...,\eta]}$ and $\texttt{msk}=\pmb{s}$.
    It returns the pair $(\texttt{pp}, \texttt{msk})$.
    
    \item $\mathcal{E}_{\text{SIFE}}.\text{DKGen}(\texttt{msk}, \pmb{y})$: The algorithm outputs the function derived key $\text{dk}_{\pmb{y}}=\langle\pmb{y},\pmb{s}\rangle$ on the inputs of master secret key $\texttt{msk}$ and vector $\pmb{y}$.
    
    \item $\mathcal{E}_{\text{SIFE}}.\text{Enc}(\texttt{pp}, \pmb{x})$: The algorithm first chooses a random $r\sample \ZZ_{p}$ and computes $\textit{ct}_0 = g^{r}$. For each $i \in [1, ..., \eta]$, it computes $\textit{ct}_i = h_i^r\cdot g^{x_{i}}$.
    Then the algorithm outputs the ciphertext $\textit{ct} = (\textit{ct}_0, \{\textit{ct}_i\}_{i \in [1, ..., \eta]})$.
    
    \item $\mathcal{E}_{\text{SIFE}}.\text{Dec}(\texttt{pp}, ct, \texttt{dk}_{\pmb{y}}, \pmb{y})$: The algorithm takes the ciphertext \textit{ct}, the public key $\texttt{msk}$ and functional key $\texttt{dk}_{\pmb{y}}$ for the vector $\pmb{y}$, and returns the discrete logarithm in basis $g$, i.e., $g^{\langle\pmb{x},\pmb{y}\rangle} = \prod_{i\in[1, ..., \eta]}\textit{ct}_{i}^{y_{i}}/ \textit{ct}_{0}^{\texttt{dk}_{f}}$.
\end{itemize}

\subsection{Multi-input FEIP Construction}

The multi-input functional encryption scheme for the inner-product $f_{\text{MIIP}}((\pmb{x}_1, ..., \pmb{x}_n), \pmb{y})$ is defined as  follows:
\begin{equation*}
\begin{split}
    \mathcal{E}_{\text{MIFE}}=(\mathcal{E}_{\text{MIFE}}.\text{Setup}, \mathcal{E}_{\text{MIFE}}.\text{SKDist}, \mathcal{E}_{\text{MIFE}}.\text{DKGen}, \\ \mathcal{E}_{\text{MIFE}}.\text{Enc}, \mathcal{E}_{\text{MIFE}}.\text{Dec})
\end{split}
\end{equation*}
The specific construction of each algorithm is as follows:
\begin{itemize}
    \item $\mathcal{E}_{\text{MIFE}}.\text{Setup}(1^{\lambda}, \vec{\eta}, n)$:
    The algorithm first generates secure parameters as
    $\mathcal{G}=(\mathbb{G}, p, g)$ $\sample$ $\textit{GroupGen}(1^{\lambda})$, and then generates several samples as $a \sample \mathbb{Z}_{p}$, $\pmb{a}=(1,a)^{\intercal}$, $\forall i \in [1,...,n]:$     $\pmb{W}_{i} \sample \mathbb{Z}^{\eta_{i}\times 2}_{p}$, $\pmb{u}_{i} \sample \mathbb{Z}^{\eta_{i}}_{p}$.
    Then, it generates the \textit{master public key} and \textit{master private key} as $\texttt{mpk} = (\mathcal{G}, g^{\pmb{a}}, g^{\pmb{Wa}})$, 
    $\texttt{msk} = (\pmb{W}, (\pmb{u}_{i})_{i \in [1,...,n]})$.
    
    \item $\mathcal{E}_{\text{MIFE}}.\texttt{SKDist}(\texttt{mpk}, \texttt{msk}, \textit{id}_{i})$:
    It looks up the existing keys via $\textit{id}_{i}$ and returns the \textit{party secret key} as $\texttt{sk}_{i} = (\mathcal{G}, g^{\pmb{a}}, (\pmb{Wa})_{i}, \pmb{u}_{i})$.
    
    \item $\mathcal{E}_{\text{MIFE}}.\textit{DKGen}(\texttt{mpk}, \texttt{msk}, \pmb{y})$:
    The algorithm first partitions $\pmb{y}$ into $(\pmb{y}_1||\pmb{y}_{2}||...||\pmb{y}_{n})$, where $|\pmb{y}_{i}|$ is equal to $\eta_{i}$.
    Then it generates the function derived key as follows:
    $\texttt{dk}_{f,\pmb{y}} = (\{\pmb{d}^{\intercal}_{i} \leftarrow \pmb{y}^{\intercal}_{i}\pmb{W}_{i}\}, z \leftarrow \sum{\pmb{y}^{\intercal}_{i}\pmb{u}_{i}})$. 
    
    \item $\mathcal{E}_{\text{MIFE}}.\texttt{Enc}(\texttt{sk}_{i}, \pmb{x}_{i})$:
    The algorithm first generates a random nonce $r_{i} \leftarrow_{R} \mathbb{Z}_{p}$, and then computes the ciphertext as follows:
    $\pmb{ct}_{i} = (\pmb{t}_{i} \leftarrow g^{\pmb{a}r_{i}}, \pmb{c}_{i} \leftarrow g^{\pmb{x}_{i}} g^{\pmb{u}_{i}} g^{(\pmb{Wa})_{i}r_{i}})$.
    
    \item $\mathcal{E}_{\text{MIFE}}.\texttt{Dec}(\pmb{ct}, \texttt{dk}_{f,\pmb{y}})$: 
    The algorithm first calculates as follows:
    $C = \frac{\prod_{i \in [1, ..., n]}([\pmb{y}^{\intercal}_{i}\pmb{c}_{i}]/[\pmb{d}^{\intercal}_{i}\pmb{t}_{i}])}{z}$,
    and then recovers the function result as 
    $f((\pmb{x}_{1}, \pmb{x}_{2}, ..., \pmb{x}_{n}),\pmb{y}) = \log_{g}(C)$.
\end{itemize}

\end{document}